\documentclass[pdflatex,iicol,Numbered]{sn-jnl}

\usepackage{graphicx}%
\usepackage{multirow}%
\usepackage{amsmath,amssymb,amsfonts}%
\usepackage{amsthm}%
\usepackage{mathrsfs}%
\usepackage[title]{appendix}%
\usepackage{xcolor}%
\usepackage{textcomp}%
\usepackage{manyfoot}%
\usepackage{algorithm}%
\usepackage{algorithmicx}%
\usepackage{algpseudocode}%
\usepackage{listings}%
\usepackage{array}
\usepackage{lmodern}
\usepackage{arydshln}
\usepackage{hyperref}
\usepackage[percent]{overpic}
\usepackage{subcaption}
\usepackage{float}
\usepackage[shortlabels]{enumitem}
\usepackage{tabularx}
\usepackage{academicons}
\usepackage{booktabs}  
\usepackage{siunitx}

\theoremstyle{thmstyleone}%
\theoremstyle{thmstyletwo}%

\theoremstyle{thmstylethree}%

\begin{document}

\title[YolovN-CBi: A Lightweight and Efficient Architecture for Real-Time Detection of Small UAVs]{YolovN-CBi: A Lightweight and Efficient Architecture for Real-Time Detection of Small UAVs}


\author*[1]{\fnm{Ami Pandat} \sur{Author}}\email{a.pandat05@gmail.com}

\author[2]{\fnm{Punna Rajasekhar} \sur{Author}}\email{rajs@barc.gov.in}
\equalcont{These authors contributed equally to this work.}

\author[1,2]{\fnm{Gopika Vinod} \sur{Author}}

\author[1,2]{\fnm{Rohit Shukla} \sur{Author}}

\affil*[1]{\orgdiv{Homi Bhabha National Institute},  \orgaddress{\city{Mumbai}, \postcode{400094}, \state{Maharashtra}, \country{India}}}

\affil[2]{\orgdiv{Bhabha Atomic Research Center}, \orgaddress{\city{Mumbai}, \postcode{400094}, \state{Maharashtra}, \country{India}}}


\abstract{Unmanned Aerial Vehicles, commonly known as, drones pose increasing risks in civilian and defense settings, demanding accurate and real-time drone detection systems. However, detecting drones is challenging because of their small size, rapid movement, and low visual contrast. A modified architecture of YolovN  called the YolovN-CBi is proposed that incorporates the Convolutional Block Attention Module (CBAM) and the Bidirectional Feature Pyramid Network (BiFPN) to improve sensitivity to small object detections. A curated training dataset consisting of 28K images is created with various flying objects and a local test dataset is collected with 2500 images consisting of very small drone objects. The proposed architecture is evaluated on four benchmark datasets, along with the local test dataset. The baseline Yolov5 and the proposed Yolov5-CBi architecture outperform newer Yolo versions, including Yolov8 and Yolov12, in the speed–accuracy trade-off for small object detection. Four other variants of the proposed CBi architecture are also proposed and evaluated, which vary in the placement and usage of CBAM and BiFPN. These variants are further distilled using knowledge distillation techniques for edge deployment, using a Yolov5m-CBi teacher and a Yolov5n-CBi student. The distilled model achieved a mAP@0.5:0.9 of 0.6573, representing a 6.51\% improvement over the teacher's score of 0.6171, highlighting the effectiveness of the distillation process. The distilled model is 82.9\% faster than the baseline model, making it more suitable for real-time drone detection. These findings highlight the effectiveness of the proposed CBi architecture, together with the distilled lightweight models in advancing efficient and accurate real-time detection of small UAVs.}

\keywords{Aerial Surveillance, CBAM, Drone Detection, Knowledge Distillation}



\maketitle

\section{Introduction}\label{intro}
The proliferation of Unmanned Aerial Vehicles (UAVs), commonly known as drones, has significantly expanded in recent years due to their versatility and cost-effectiveness. Drones are now widely used in diverse fields such as agriculture for crop monitoring and spraying, aerial photography, surveillance, infrastructure inspections, environmental monitoring, disaster management, and military operations. Although drones have substantially simplified complex tasks and have increased productivity in civilian contexts, they also pose significant risks, particularly in military, security, and sensitive areas \cite{4,19,53}. Recent conflicts and security incidents have illustrated the potential for malicious drone activity, such as unauthorized surveillance, unauthorized data collection, and weaponized payload delivery. These evolving threats underscore the urgent need for effective drone detection and countermeasure systems to protect sensitive infrastructure, ensure airspace safety, and mitigate associated risks \cite{76}.

During the past decade, researchers have proposed numerous solutions to detect drones in airspace\cite{77}, yet this remains a complex and challenging task due to the small size, rapid movement, and often unpredictable trajectories of drones. Although traditional detection methods, such as radar, RF analysis, and acoustics, have shown some success, they often struggle with accuracy and adaptability in dynamic environments. In recent years, computer vision-based approaches have gained prominence as an alternative for civilian and low-cost applications \cite{75}, especially after the success of deep learning and Convolutional Neural Networks (CNNs) for image and video analysis. Various deep learning models are proposed for object detection tasks, viz. Faster R-CNN \cite{54}, Region-Based Fully Convolutional Networks (R-FCN) \cite{78}, You Only Look Once (Yolo) \cite{61}, Single Shot Detector (SSD) \cite{55} etc. Among these, Yolo and its versions (v3, v5, v9, v12, etc.) achieve a better speed-accuracy trade-off which is required for real-time applications, such as drone detection.

The state-of-the-art Yolo versions are deeper and more accurate but suffer from high inference time. A lightweight yet high-performance variant is proposed to reduce inference time, named YolovN-CBi (where N represents the Yolo version, N = 5 to 12). The proposed variant augments the YolovN backbone with the Convolutional Block Attention Module (CBAM) \cite{66} and the Bidirectional Feature Pyramid Network (BiFPN) \cite{94} to enhance the ability of the model to focus on relevant spatial and channel-wise information while improving multiscale feature aggregation, which is particularly beneficial for detecting small drones at long distances. The proposed YolovN-CBi variant yields faster inference than the higher version models like Yolov9 through Yolov12, but with comparable mean average precision (mAP). 

The Yolov5m-CBi model is distilled for real-time deployment on edge devices. Knowledge Distillation (KD) \cite{74} is applied where a smaller and faster student model learns from a pretrained teacher model (Yolov5-CBi). Various KD strategies are evaluated for distillation and it is observed that the Adversarial Knowledge Distillation (AKD) model achieves the best trade-off with an inference time of 2.4 ms, which is 82.9\% improvement over the teacher model. Notably, it also surpassed the teacher in detection accuracy, achieving a mAP@0.5 of 0.9902 compared to the teacher's 0.9741, while maintaining high precision and recall. The distilled model also outperforms state-of-the-art drone detection models such as LA-YOLO\cite{97}, SEB-YOLOv8s\cite{99},DRBD-YOLOv8s\cite{98},IASL-YOLO\cite{95}. These results demonstrate that KD enables compact models to deliver high-performance drone detection, making them well-suited for time-sensitive and resource-constrained surveillance systems.

\subsection*{Contributions}

The main contributions of this work are summarized as follows:

\begin{itemize}
    \item \textbf{Comprehensive Benchmarking.} A systematic evaluation of multiple versions of Yolo-based drone detection models (Yolov5 to Yolov12) is presented. Performance metrics such as mAP, precision, recall, and inference speed on full HD (1920$\times$1080) resolution inputs are reported on four benchmark drone datasets and one local test dataset. This is the first unified comparison that specifically targets small drone detection.

    \item \textbf{YolovN-CBi Architecture.} An enhanced YolovN architecture that incorporates a Convolutional Block Attention Module (CBAM) into the backbone and modifies the PANet neck with a Bidirectional Feature Pyramid Network (BiFPN) is proposed for small object detection. CBAM improves spatial and channel-wise focus, while BiFPN enables efficient multiscale feature fusion. This architecture increases detection recall, particularly for drones as small as 20 pixels.

    \item \textbf{Knowledge-Distilled Student Model.} Three knowledge distillation (KD) strategies are evaluated on both standard and modified teacher-student configurations. The proposed Yolov5m-CBi is distilled into a lightweight student model using adversarial distillation, which is \textbf{82.9\% faster} than the teacher while maintaining comparative accuracy, allowing deployment on resource-constrained edge devices.
\end{itemize}

The remainder of this paper is organized as follows. Section \ref{sec:related_work} reviews related work on drone detection, Yolo-based object detectors, and knowledge distillation techniques. Section \ref{sec:methods} summarizes various modules, and Section \ref{sec:methodology} details the proposed YolovN-CBi architecture. Section \ref{sec:dataset} introduces the benchmark drone datasets used for training and evaluation. Section \ref{sec:experiments} presents a comprehensive performance comparison between Yolo versions and the proposed models. Section \ref{sec:kd} discusses various knowledge distillation strategies and their results, highlighting the most suitable student model for real-time small drone detection on edge devices. Section \ref{sec:comparison} presents the comparison of the proposed model with other state-of-the-art models on a benchmark dataset. Finally, Section \ref{sec:conclusion} concludes the paper and outlines the directions for future work.

\section{Related Work}\label{sec:related_work}

This section reviews recent advances in drone detection methods, with a focus on various Yolo models and their enhancements using attention mechanisms. It also summarizes techniques for deploying large models on edge devices using knowledge distillation.

\subsection{Small Drone Detection}\label{small drone lit}

Deep learning techniques have significantly advanced drone detection, based on Convolutional Neural Networks, such as Yolo, and Faster R-CNN. Of these, Yolo models provide real-time detection capabilities essential for rapid response applications \cite{43,26,24}. Faster R-CNN \cite{54} and SSD \cite{55} are known for their high precision and multiscale detection, and Mask R-CNN \cite{56} extends this with pixel-level segmentation. Yolo has gained attention for drone detection due to its balance between speed and accuracy \cite{57}. Early Yolov3 implementations have shown effectiveness in detecting drones in real time \cite{58,59}. However, Yolov3 faced limitations in identifying small or distant drones under challenging conditions such as low visibility or adverse weather. These drawbacks are addressed by Yolov4 by introducing architectural and optimization modifications that improve semantic feature extraction \cite{60}. The release of Yolov5 in 2020 \cite{61} marked a significant advance with a streamlined architecture and better training pipelines, leading to improved detection performance. It includes five variants (n, s, m, l and xl), offering trade-offs between speed and accuracy. Yolov5s is optimized for real-time use, while Yolov5s6 offers increased accuracy for complex scenes. 

Following Yolov5, a series of newer Yolo versions were introduced with progressive improvements in object detection performance, efficiency, and adaptability. The key advances in Yolov6 through Yolov12 are summarized as follows:

\begin{itemize}
\item \textbf{Yolov6} \cite{71} improved training efficiency through an anchor-free detection head, removing the need for manual anchor design and simplifying model optimization.

\item \textbf{Yolov7} \cite{88} introduced Extended Efficient Layer Aggregation Networks (E-ELAN) and implicit label assignment, allowing deeper and more accurate models without significant computational overhead.

\item \textbf{Yolov8} \cite{87} emphasized usability and deployment flexibility, integrating features like native instance segmentation support and mosaic data augmentation

\item \textbf{Yolov9} \cite{9} enhanced robustness in visually complex environments through dynamic feature fusion and improved normalization layers, targeting generalization in diverse drone detection contexts.

\item \textbf{Yolov10} \cite{86} adopts adaptive scaling strategies, allowing the model to dynamically adjust its depth and width to better handle objects at varying scales and resolutions.

\item \textbf{Yolov11} \cite{85} is focused on edge deployment and optimized for low-power devices using lightweight backbones, efficient convolutional blocks, and aggressive pruning strategies.

\item \textbf{Yolov12} \cite{84} marked a shift towards attention-based architectures while maintaining real-time performance.

\end{itemize}

Despite Yolov5 and later versions providing excellent real-time performance, small object detection still remains a notable weakness, particularly in aerial imagery where targets occupy only a few pixels \cite{18}. Although state-of-the-art vision-based detection \cite{53} achieves a high mean Average Precision (mAP) of 0.99, it exhibits notable limitations in detecting small drones, especially when the target is located far from the camera. In such scenarios, the drone typically occupies only 15–20 pixels in the image, resulting in limited visual information. This significantly limits the system’s ability to reliably detect the drone. This issue arises from aggressive downsampling and low feature resolution in deeper layers. Several enhancements and modifications to Yolo for small drone detection are proposed to address this, such as Javan et al. \cite{60} adopt the pruning of Yolov4 layers and uses the copy–paste data augmentation technique to achieve 90.5\% mAP with a 60.4\% reduction in inference time outperforming RetinaNet, SSD, and Faster R-CNN on their small drone dataset. Aydin and Singha \cite{18} trained Yolov5 on 2400 drone/bird images, achieving 90.4\% mAP@0.5, an improvement of 21.6\% over Yolov4, along with precision 0.912 and recall 0.804. Zhao et al. \cite{62} proposed TGC-Yolov5, which integrates transformer and attention mechanisms, and achieved 0.848 AP on SUAV-DATA, which is 2.5\% higher than Yolov5 baseline. Liu et al. \cite{63} developed GL-YOMO by combining Yolov5 with motion analysis, achieving 0.82 AP at 21.6 FPS. However, these approaches often assume clear sky backgrounds and favorable lighting. 

Later versions of YOLO, from YOLOv8 to YOLOv12, have introduced task-specific architectural modifications aimed at improving YOLOv5. Although newer versions of YOLO (v6 to v12) typically achieve higher mAP scores on benchmarks, viz. COCO \cite{81}, they often involve greater architectural complexity, higher computational demands, and can show inconsistent performance in specific applications. Several state-of-the-art studies have highlighted that the newer YOLO versions do not outperform YOLOv5, particularly in specialized domains. Khanam et al. reported that for solar panel defect detection, the YOLOv5s model achieved a higher precision and a shorter inference time of 7.1 ms/frame compared to YOLOv8s for a full HD input image size \cite{65}. 

Jegham et al. have conducted a detailed study of all new Yolo versions for two datasets for different applications and found that YOLOv12vs incorporates attention-centric modules and achieves a high mAP of 47.6\% , yet performs poorly in specific tasks \cite{51}, suggesting that accuracy improvements do not always justify the additional computational cost. In contrast, YOLOv11 offers a better balance between performance and efficiency. Similarly, YOLOv9 and YOLOv10 demonstrate specific trade-offs: YOLOv9 shows strong precision but struggles with detecting small objects, while YOLOv10 emphasizes speed but compromises accuracy in scenarios with overlapping objects. Hence, YOLOv7 through YOLOv12 focus on general-purpose object detection improvements rather than tuning for domain-specific challenges such as drone detection.

In contrast, the YOLOv5s variant continues to serve as a baseline for accurate drone detection in real time. Ultralytics emphasizes YOLOv5’s optimization for speed, particularly in real-time applications \cite{64}, reinforcing the notion that newer is not always better especially in tasks requiring low-latency and high-precision performance, such as small drone detection.

Using Yolov5 to detect small drones against cluttered backgrounds (including birds) is particularly challenging for vanilla CNN detectors. Studies show that CNNs tend to confuse drones with birds due to similar size and motion, yielding high false positives. Recent works emphasize on attention mechanisms and also usage of multi-frame cues. Cazzato et al. \cite{50} applied spatio-temporal attention to separate drones from birds, reducing the false-positive rate of Yolov5 by 20\%.  CBAM (Convolutional Block Attention Module) \cite{67} is a lightweight attention block that sequentially refines feature maps via channel and spatial attention. A recent survey \cite{92} for small object detection highlights the integration of CBAM and Bidirectional feature pyramid network, based on application. Inspired by this research, the proposed Yolov5-CBi integrates CBAM and BiFPN for a better small object detection task. 

\subsection{Knowledge Distillation Strategies for Small Drone Detection}\label{kd-lit}

Knowledge distillation (KD) is a strategy to improve the performance of less complex models by transferring knowledge from a larger, more accurate and pre-trained teacher model \cite{74}. Unlike classification-based distillation, object detection distillation must handle both classification and localization outputs, along with more complex feature hierarchies and anchor behaviors. \cite{68}.

Distillation strategies in object detection generally fall into three categories: (1) \textit{Logit distillation}, where the student mimics the teacher’s class confidence scores and bounding box outputs; (2) \textit{Feature distillation}, where intermediate feature maps from the backbone or neck are aligned between the teacher and the student; and (3) \textit{Instance-aware distillation}, which emphasizes spatial regions with high semantic importance, such as object centers or edges \cite{93}. 

Recent work such as LT-DETR \cite{72} has shown that dual-level KD (logit + feature) can distill a detector by up to 50\% without accuracy loss. Similarly, Ultralytics' Yolo-NAS (2023) integrates KD into the training pipeline, transferring both semantic and structural knowledge from a larger teacher model to improve student’s performance. Most Yolo distillation studies follow a homogeneous teacher–student setup, where a larger Yolo variant serves as the teacher for a smaller counterpart \cite{69} as architectural consistency simplifies feature alignment. In contrast, heterogeneous distillation methods that use two-stage or transformer-based teachers for Yolo students have been explored in specialized domains \cite{70}. Self-distillation techniques are proposed in Yolov6, which uses an adaptive method where the model learns from an earlier exponential moving average (EMA) version of itself \cite{71}. 

Several studies improve the performance of KD by integrating attention mechanisms. Cheng et al. \cite{73} proposed a Global Attention Mechanism Fusion (GAMF) module in Yolov5 to capture both global and local contexts. Jobaer et al. \cite{69} used Convolutional Block Attention Modules (CBAM) in the neck of teacher model to refine its features, improving the student’s ability to mimic meaningful activations. Their framework also used self-supervised learning with blurry and deblurred image pairs, allowing the student to learn robust features from raw inputs.

\section{Key Components}\label{sec:methods}

This section introduces the key architectural components used in the proposed model design: the Convolutional Block Attention Module (CBAM) and a BiFPN-inspired multiscale feature fusion mechanism. These modules are adopted within the YOLOv5 framework to enhance multiscale feature aggregation and improve attention to salient regions in the input image required for small object detection.

\subsection{Convolutional Block Attention Module (CBAM)}

\begin{figure*}
    \centering
    \includegraphics[width=0.7\linewidth]{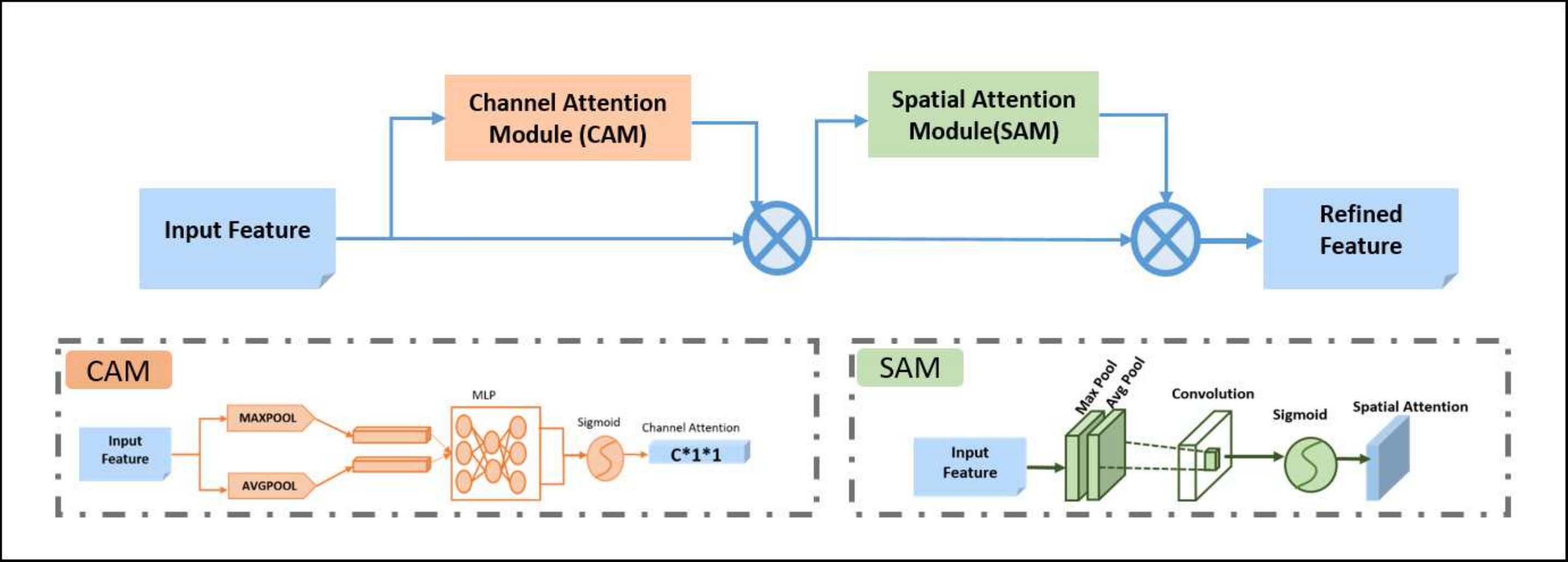}
    \caption{Convolutional Bottleneck Attention Module (CBAM) architecture}
    \label{fig:cbam}
\end{figure*}

CBAM \cite{66} is an end-to-end trainable lightweight attention module that can be applied to any feature map of a deep neural network to improve their representational power. CBAM consists of two modules, the Channel Attention Module (CAM) and the Spatial Attention Module (SAM), which computes attention maps along the channel and spatial dimensions, respectively, and then multiplies these attention maps with the input feature to refine it as shown in Eq.\ref{eq:cbam_final}. The architecture of CBAM is presented in Figure \ref{fig:cbam}.

\begin{equation}
F'' = M_s(M_c(F) \otimes F) \otimes (M_c(F) \otimes F)
\label{eq:cbam_final}
\end{equation}

where:
\begin{itemize}
    \item \( F \) is the input feature map,
    \item \( M_c(F) \) is the channel attention map (\(C \times 1 \times 1\)),
    \item \( M_s(F') \) is the spatial attention map (\(1 \times H \times W\)),
    \item \( \otimes \) denotes element-wise multiplication (with broadcasting).
\end{itemize}

\subsubsection{Channel Attention Module (CAM)}

The Channel Attention Module (CAM) is designed to focus on the most informative channels of the input feature. Given an input feature $\mathbf{F} \in \mathbb{R}^{C \times H \times W}$, where $C$, $H$, and $W$ represent the number of channels, height, and width of the feature map, respectively, CAM first applies both average pooling and max pooling operations along the spatial dimension, generating two different context descriptors: $\mathbf{F}_{\text{avg}}^{\text{c}}$ and $\mathbf{F}_{\text{max}}^{\text{c}}$. These descriptors are then forwarded to a shared multi-layer perceptron (MLP) with one hidden layer, which outputs a channel attention map $\mathbf{M}_\text{c} \in \mathbb{R}^{C \times 1 \times 1}$:

\begin{equation}
   \mathbf{M}_\text{c}(\mathbf{F}) = \sigma(\text{MLP}(\text{AvgPool}(\mathbf{F})) + \text{MLP}(\text{MaxPool}(\mathbf{F}))) 
   \label{eq:cam1}
\end{equation}

where $\sigma$ denotes the sigmoid function. The resulting attention map is then multiplied by the input feature to emphasize the significant channels:

\begin{equation}
  \mathbf{F}' = \mathbf{M}_\text{c}(\mathbf{F}) \otimes \mathbf{F}  
  \label{eq:cam2}
\end{equation}

where $\otimes$ denotes element-wise multiplication.

\subsubsection{Spatial Attention Module (SAM)}

The refined feature $\mathbf{F}'$ of CAM is fed into the Spatial Attention Module (SAM) to focus on the most informative regions in the spatial domain. SAM applies both average pooling and max-pooling along the channel dimension, generating two 2D maps: $\mathbf{F}_{\text{avg}}^{\text{s}}$ and $\mathbf{F}_{\text{max}}^{\text{s}}$. These maps are then concatenated and convolved with a standard convolution operation to produce the spatial attention map $\mathbf{M}_\text{s} \in \mathbb{R}^{1 \times H \times W}$:

\begin{equation}
    \mathbf{M}_\text{s}(\mathbf{F'}) = \sigma(\text{Conv}([\text{AvgPool}(\mathbf{F'}) \, : \, \text{MaxPool}(\mathbf{F'})]))
    \label{eq:sam2}
\end{equation}

where $[ \, : \, ]$ represents the concatenation along the channel axis. The resulting attention map is then multiplied by the refined feature to emphasize the significant spatial locations:

\[
\mathbf{F}'' = \mathbf{M}_\text{s}(\mathbf{F'}) \otimes \mathbf{F'}.
\]

In contrast to heavier attention mechanisms such as DETR \cite{100}, which introduces significant overhead through global self-attention, or SimAM \cite{83}, which estimates the importance of neuron without explicitly modeling spatial information, CBAM provides a lightweight yet effective attention mechanism. Its dual focus on spatial and channel-wise attention helps the model better localize small targets without compromising speed, an essential requirement for drone detection systems operating in edge computing environments. In the proposed model, CBAM is introduced into the backbone and neck of the YOLOv5 architecture. 

\subsection{BiFPN-inspired Feature Fusion}
\begin{figure*}
\centering
\includegraphics[width=0.7\linewidth]{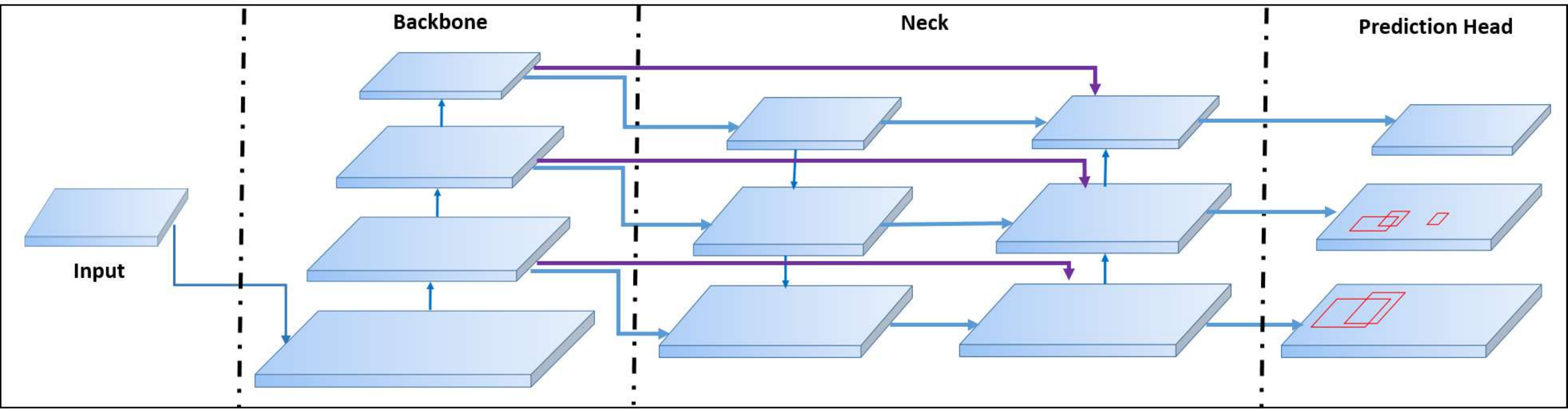}
\caption{YOLOv5 architecture with BiFPN-inspired enhancement. Purple arrows represent added connections from backbone to neck that help preserve small-scale features.}
\label{fig:bi}
\end{figure*}

In order to improve the aggregation of features on multiple scales, a fusion technique inspired by the Bidirectional Feature Pyramid Network (BiFPN) \cite{94}, as illustrated in Figure \ref{fig:bi} is introduced. Instead of learnable fusion weights,  direct connections from the backbone to the neck layers are introduced, as shown by the purple arrows in Figure \ref{fig:bi}, which allows a finer information flow and improves the detection of small objects.

Although the proposed design does not implement the full BiFPN structure with iterative bidirectional paths and learnable weights, it retains the core idea of multiscale feature fusion more effectively. This lightweight approach helps preserve features from the early layers, making the architecture better suited for small object detection without adding significant computational overhead.

\section{Proposed Model and its Variants}\label{sec:methodology}
The key components introduced in the previous section are used to build five variants of any YOLOvN architecture (N represents the version of the Yolo architecture, N = 5 represents Yolov5, N = 8 represents Yolov8 etc.) with different configurations of attention and feature fusion. Each variant combines CBAM and BiFPN modules at different locations within the network to enhance the feature representation and the detection performance for small drones. In this section, the overall integration strategy and the five proposed variants are described.

\subsection*{YolovN-CBi: CBAM in Backbone and BiFPN between Backbone and Neck}
\begin{figure*}
    \centering
    \includegraphics[width=0.7\linewidth]{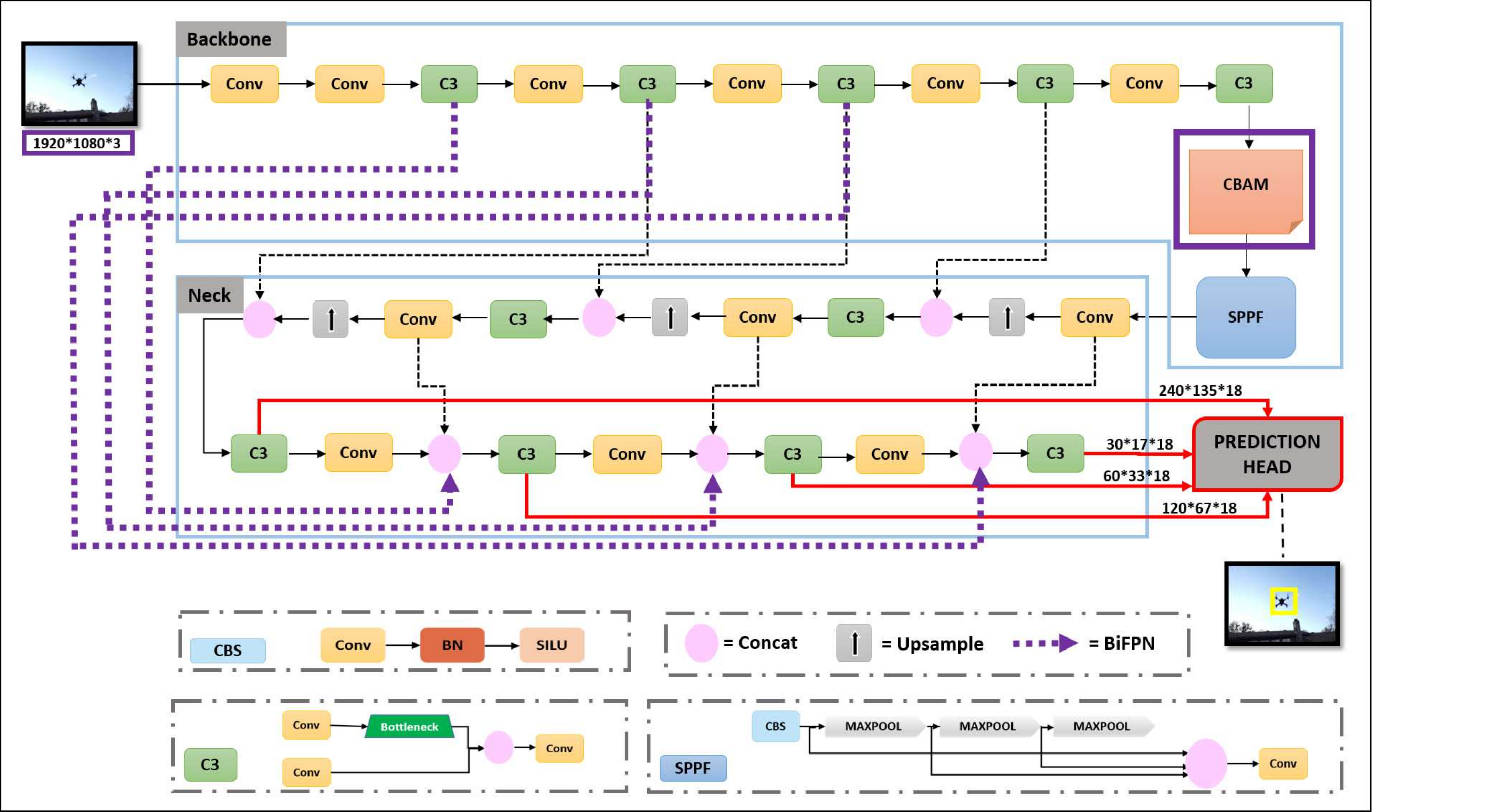}
    \caption{Yolov5-CBi architecture: Yolov5  enhanced with CBAM in backbone and BiFPN for improved attention and multiscale fusion. BiFPN integration highlighted with Purple arrows connecting backbone blocks to Neck at different scales. CBAM is placed before SPPF block in Backbone.}
    \label{fig:cbi}
\end{figure*}

In this variant, both CBAM and BiFPN-inspired fusion are introduced into the YolovN architecture. In this, CBAM is placed before the SPPF layer in the backbone and the BiFPN connections are from backbone to neck, enhancing the accuracy without significantly increasing model complexity. Figure \ref{fig:cbi} illustrates the integrated architecture for the Yolov5 version.  

\subsection*{YolovN-CBi-v2: Early Attention}
\begin{figure}
    \centering
    \includegraphics[width=0.5\linewidth]{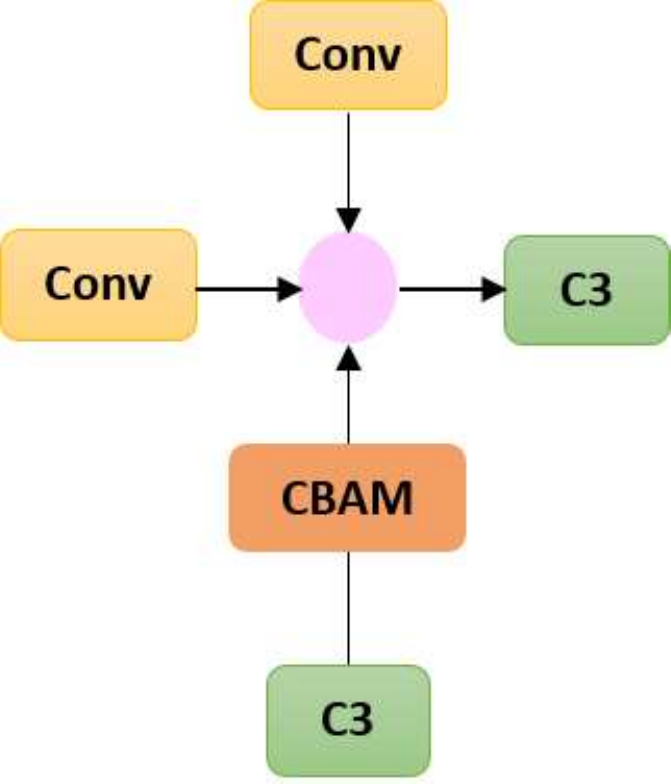}
    \caption{Early Attention: CBAM introduced before concatenating features of C3 layers from backbone}
    \label{fig:v2}
\end{figure}

In this variant, CBAM is introduced prior to concatenating the features of the C3 backbone with the features of other modules. This early application of CBAM enables the network to capture fine-grained spatial cues and focus on salient regions from the initial stages of processing. The BiFPN-style fusion in the neck then aggregates these attention-refined features across scales, preserving and enhancing their contribution to the final detection output. Figure \ref{fig:v2} illustrates the integrated architecture for the Yolov5 model.  

\subsection*{YolovN-CBi-v3: Replacing Bottleneck layers in post-concat C3 module with CBAM layers }
\begin{figure}
    \centering
    \includegraphics[width=0.5\linewidth]{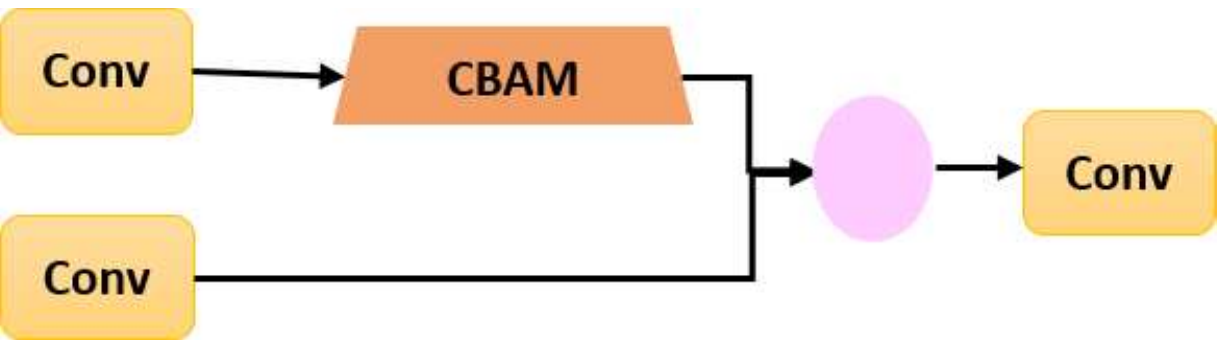}
    \caption{C3b: Modified C3 Block with CBAM replacing bottleneck blocks}
    \label{fig:c3b}
\end{figure}

\begin{figure}
    \centering
    \includegraphics[width=0.5\linewidth]{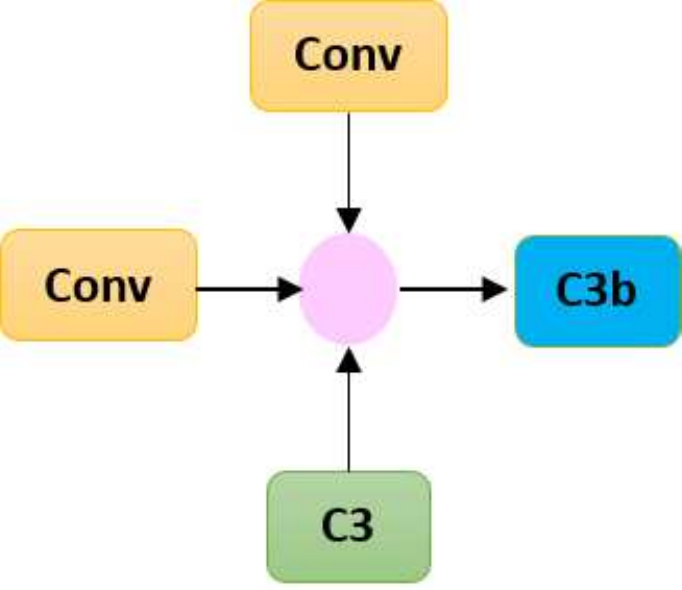}
    \caption{C3 replaced by C3b after concatenation layer}
    \label{fig:v3}
\end{figure}

Here, the standard bottleneck blocks within the C3 modules present after concatenation are replaced with CBAM, termed C3b as shown in Figure \ref{fig:c3b}. The modified concatenation structure is shown in Figure \ref{fig:v3}. This adaptation introduces attention within the aggregation stage, focusing the network on relevant features while reducing redundancy.

Bottleneck blocks, while efficient, do not have inherent attention mechanism. The proposed C3b structure uses CBAM in place of the bottleneck blocks, allows the network to refine concatenated features through spatial and channel-wise attention modules.

\subsection*{YolovN-CBi-v4: Late Attention}
\begin{figure}
    \centering
    \includegraphics[width=0.5\linewidth]{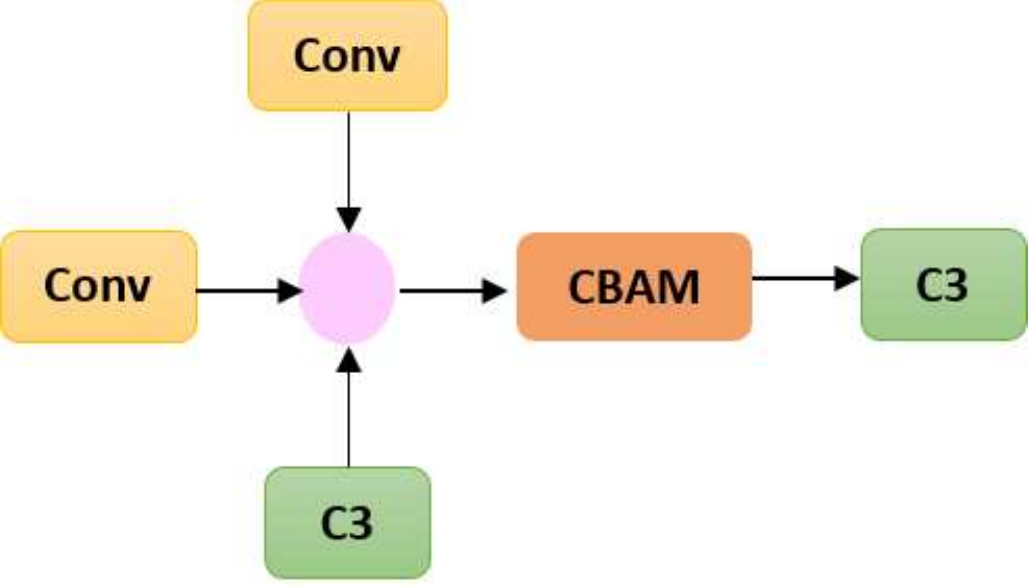}
    \caption{Late Attention: CBAM applied after concatenation}
    \label{fig:v4}
\end{figure}

Instead of modifying the bottleneck blocks of post-concat C3 layers, a CBAM module is introduced after BiFPN concatenation to focus on the aggregated multiscale features before final detection layers. Applying CBAM after feature concatenation helps the model re-emphasize critical small-object cues that may be diluted during the concatenation process. The modified concatenation structure for late attention is shown in Figure \ref{fig:v4}.

\subsection*{YolovN-CBi-v5: Replacement of all C3 blocks with C3b blocks }
\begin{figure}
    \centering
\includegraphics[width=0.5\linewidth]{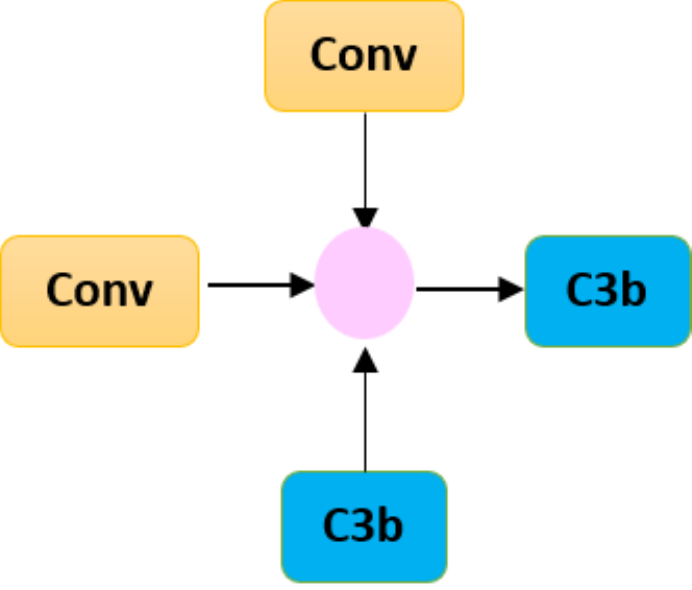}
    \caption{All C3 blocks replaced by C3b blocks}
    \label{fig:v5}
\end{figure}

In this variant, all C3 blocks are replaced by C3b blocks, effectively eliminating standard bottleneck blocks entirely. This change reduces the number of parameters while fully leveraging attention-driven refinement in the multiscale and detection stages. This is illustrated in Figure \ref{fig:v5}.

\section{Datasets}\label{sec:dataset}

A curated \textit{training} dataset called as Flying Object dataset composed of varied flying objects under different visual conditions is created as described below, and multiple publicly available datasets are used to assess model performance in diverse and real-world scenarios.

\subsection{Training Dataset: Flying Object (FO)}

A new dataset called as \textbf{Flying Object (FO)} dataset is created with various object categories, backgrounds, and lighting conditions. It combines samples from two benchmarked sources:

\begin{itemize}
    \item \textbf{Drone class (20,000 images)} - Drone images are sampled from the VisioDECT dataset \cite{79}, captured under three lighting conditions: sunny, cloudy, and evening, to simulate real-world appearance variations. It includes images of 6 types of different drones.
    
    \item \textbf{Bird and Airplane classes (8,000 images)} - Non-drone aerial objects are included using samples from the COCO dataset \cite{81} to enhance class separation and background diversity.
\end{itemize}

A small set of background images representing the deployment environment is added to the dataset to improve generalization and site-specific performance. This inclusion helps the model to adapt more effectively to local conditions during inference.

Each image is annotated with bounding boxes and a class label from the set \{\textit{drone}, \textit{bird}, \textit{airplane}\}. The dataset is divided into 90\% for training and 10\% for validation.

\subsection{Test Datasets}

Four publicly available benchmark datasets are used, along with a locally collected dataset, for a comprehensive and unbiased evaluation. Each dataset includes between 1500 and 2700 frames, featuring drones at various scales and under different environmental conditions.

\begin{itemize}
    \item \textbf{DUT-AntiUAV} \cite{80} - Captured in complex urban environments as part of the Dalian University of Technology Anti-UAV project, this dataset contains drone instances in various background settings and different scales. It consists of 2700 frames.

    \item \textbf{LRDD (Long Range Drone Detection)} \cite{49} - This dataset is designed to evaluate drone detection algorithms at extended distances, which is essential for surveillance and air traffic control applications. It contains a total of 1600 frames.

    \item \textbf{UAV-CDT} \cite{53} - A state-of-the-art dataset for object-level drone detection and tracking. It contains 1520 annotated images of three drone types, F450, Hubsan, and Phantom, captured under controlled conditions.

    \item \textbf{Anti-UAV} \cite{82} - This dataset contains video frames of drones flying under various lighting conditions, including bright sunlight, overcast skies, and low-light settings. A total of 2500 frames were extracted for testing purposes.

    \item \textbf{Local Test Dataset} - A Long-range, small-scale real-world scenario is created using a drone flying approximately 350 meters from a fixed camera and recording around 2500 images. The objects in this dataset are tiny and as small as 10 pixels, making it a difficult dataset to evaluate.
    
\end{itemize}

Together, these datasets span a broad spectrum of object scales and scene complexities. Evaluating the model using multiple independent datasets verifies robustness of the model and identifies if the model is overfitting to training data.

\begin{figure*}
    \centering
    \begin{subfigure}[t]{0.12\textwidth}
        \includegraphics[angle=270,width=\linewidth]{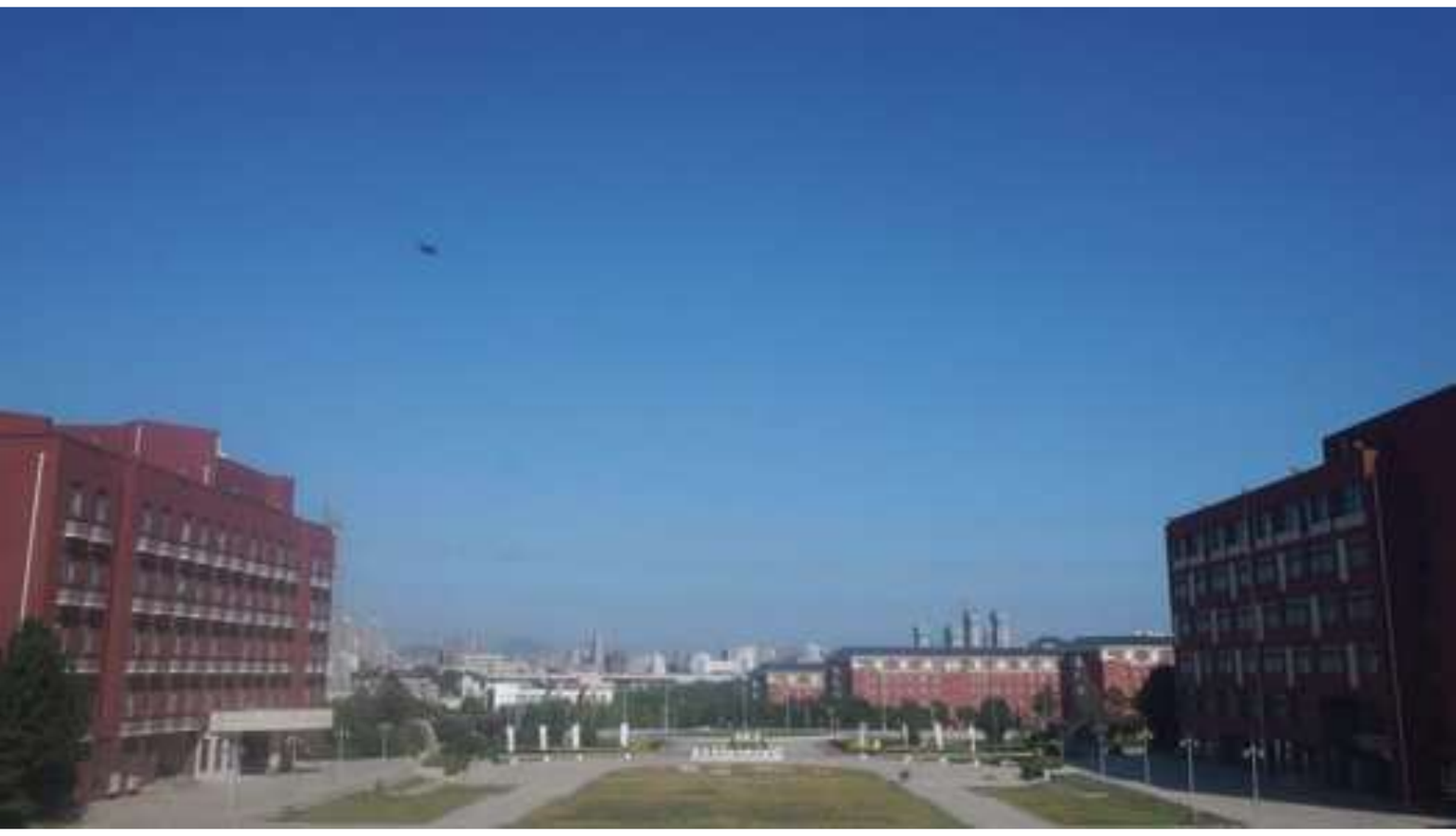}
    \end{subfigure}
    \begin{subfigure}[t]{0.12\textwidth}
        \includegraphics[angle=270,width=\linewidth]{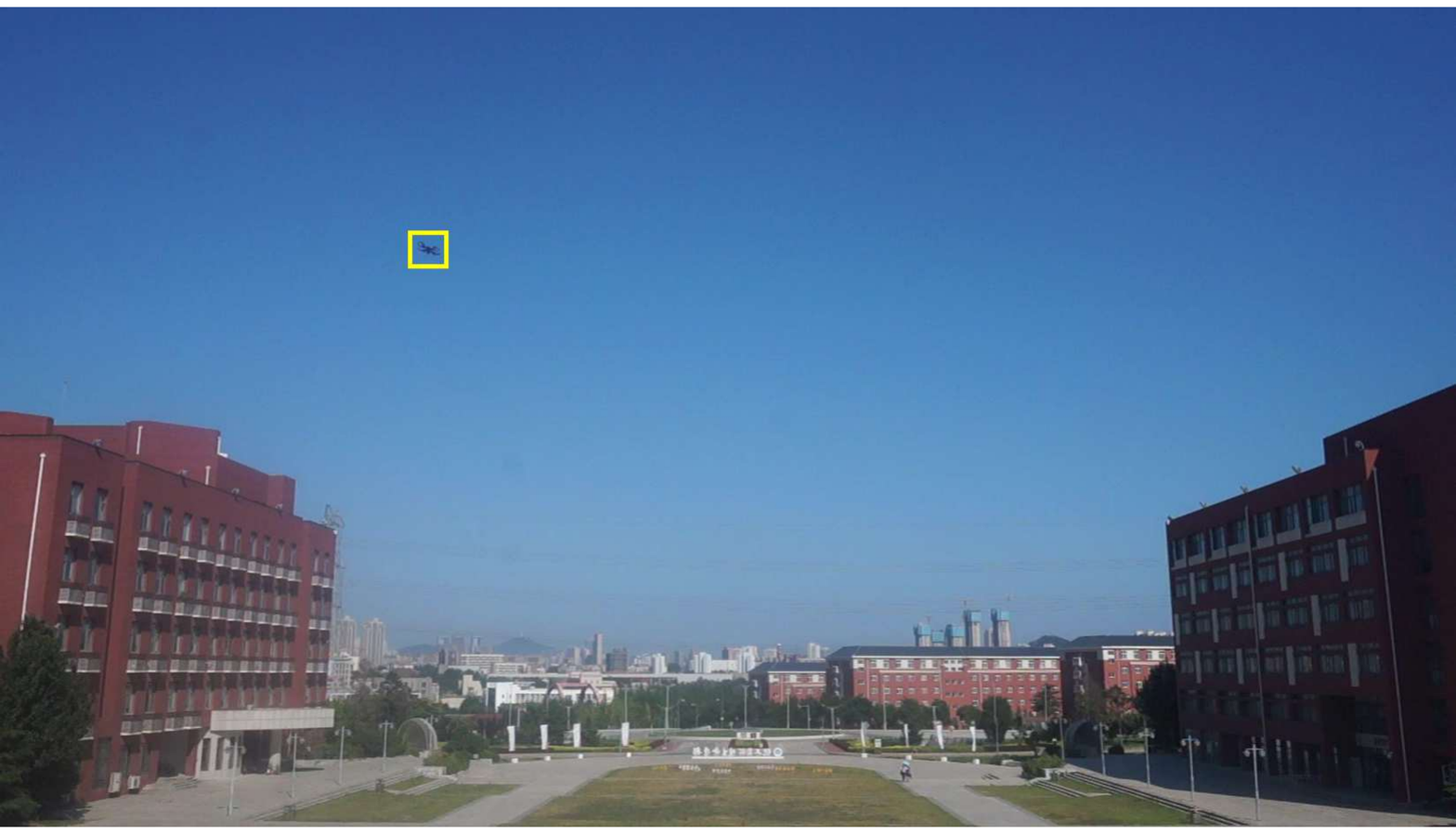}
    \end{subfigure}
    \begin{subfigure}[t]{0.12\textwidth}
        \includegraphics[angle=270,width=\linewidth]{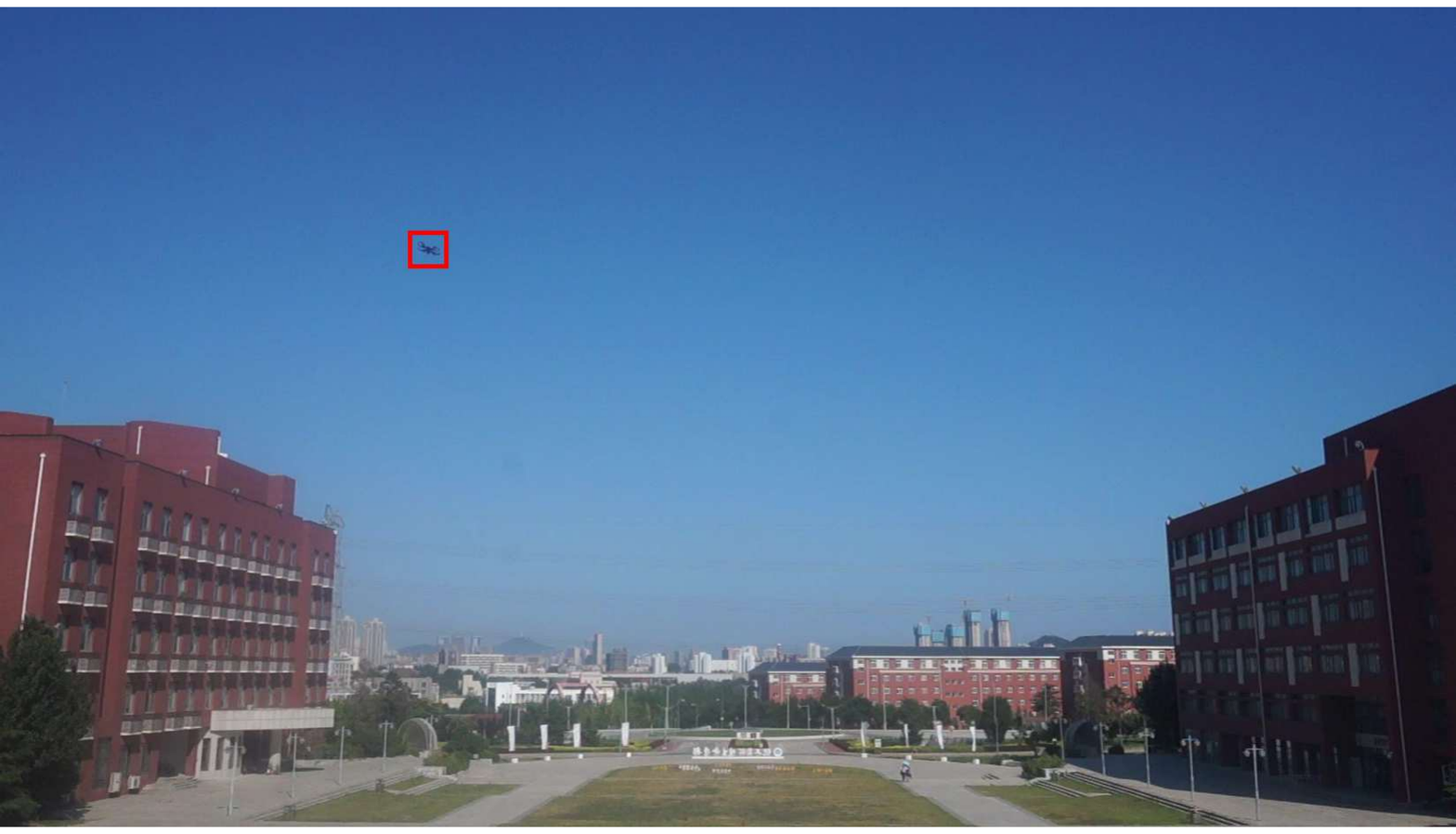}
    \end{subfigure}
    \begin{subfigure}[t]{0.12\textwidth}
        \includegraphics[angle=270,width=\linewidth]{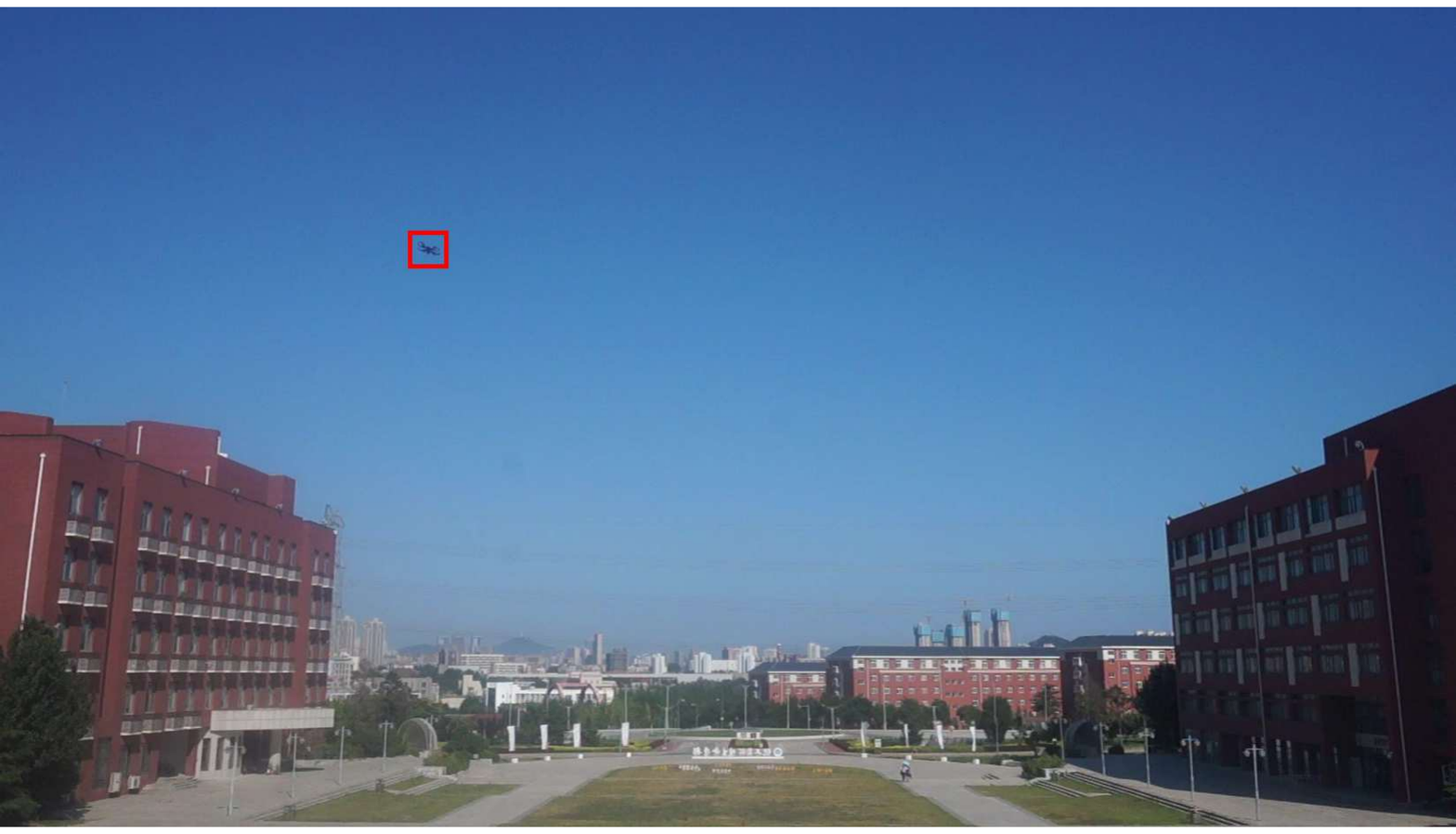}
    \end{subfigure}
    \begin{subfigure}[t]{0.12\textwidth}
        \includegraphics[angle=270,width=\linewidth]{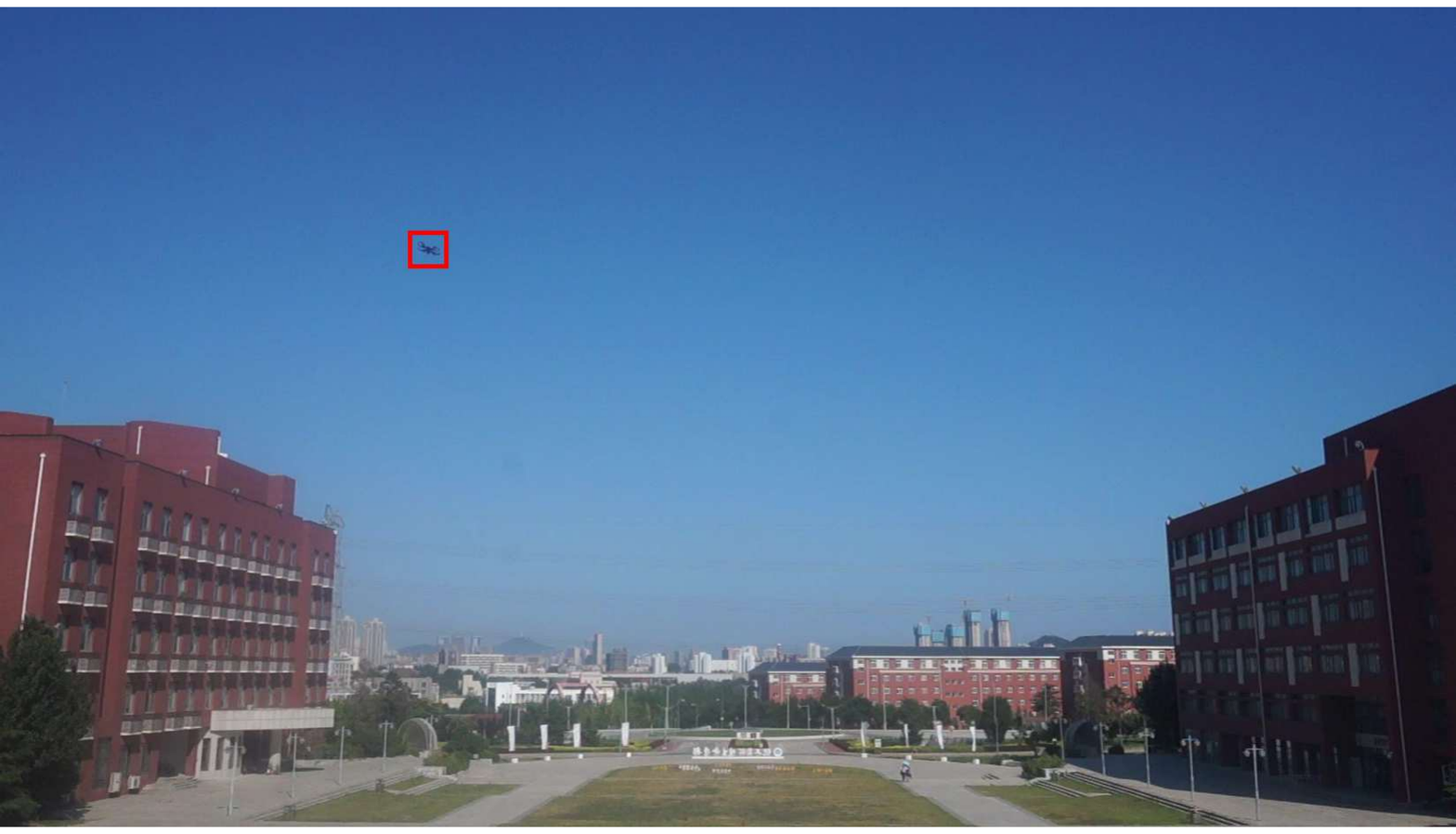}
    \end{subfigure}
    \begin{subfigure}[t]{0.12\textwidth}
        \includegraphics[angle=270,width=\linewidth]{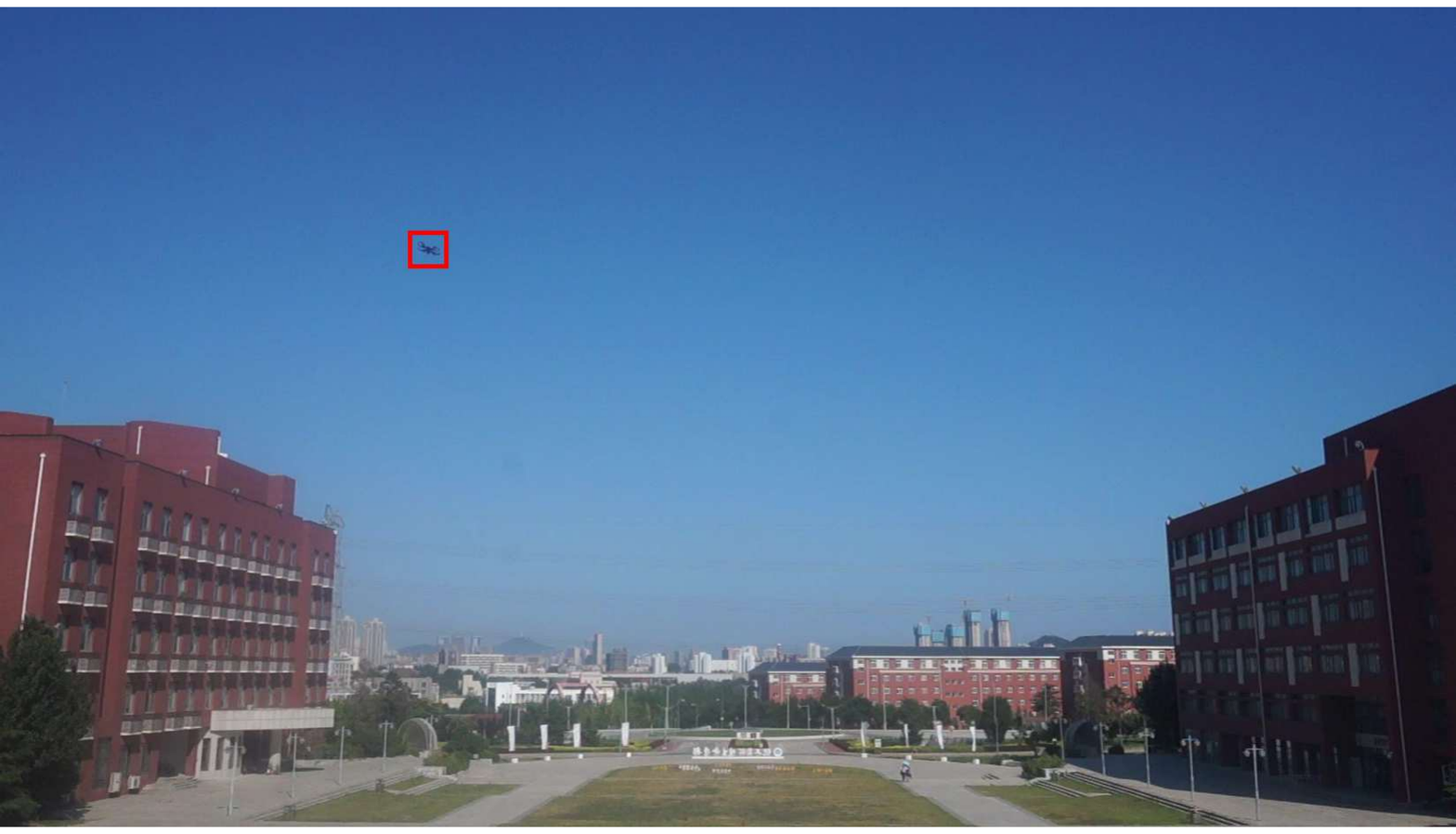}
    \end{subfigure}
    \begin{subfigure}[t]{0.12\textwidth}
        \includegraphics[angle=270,width=\linewidth]{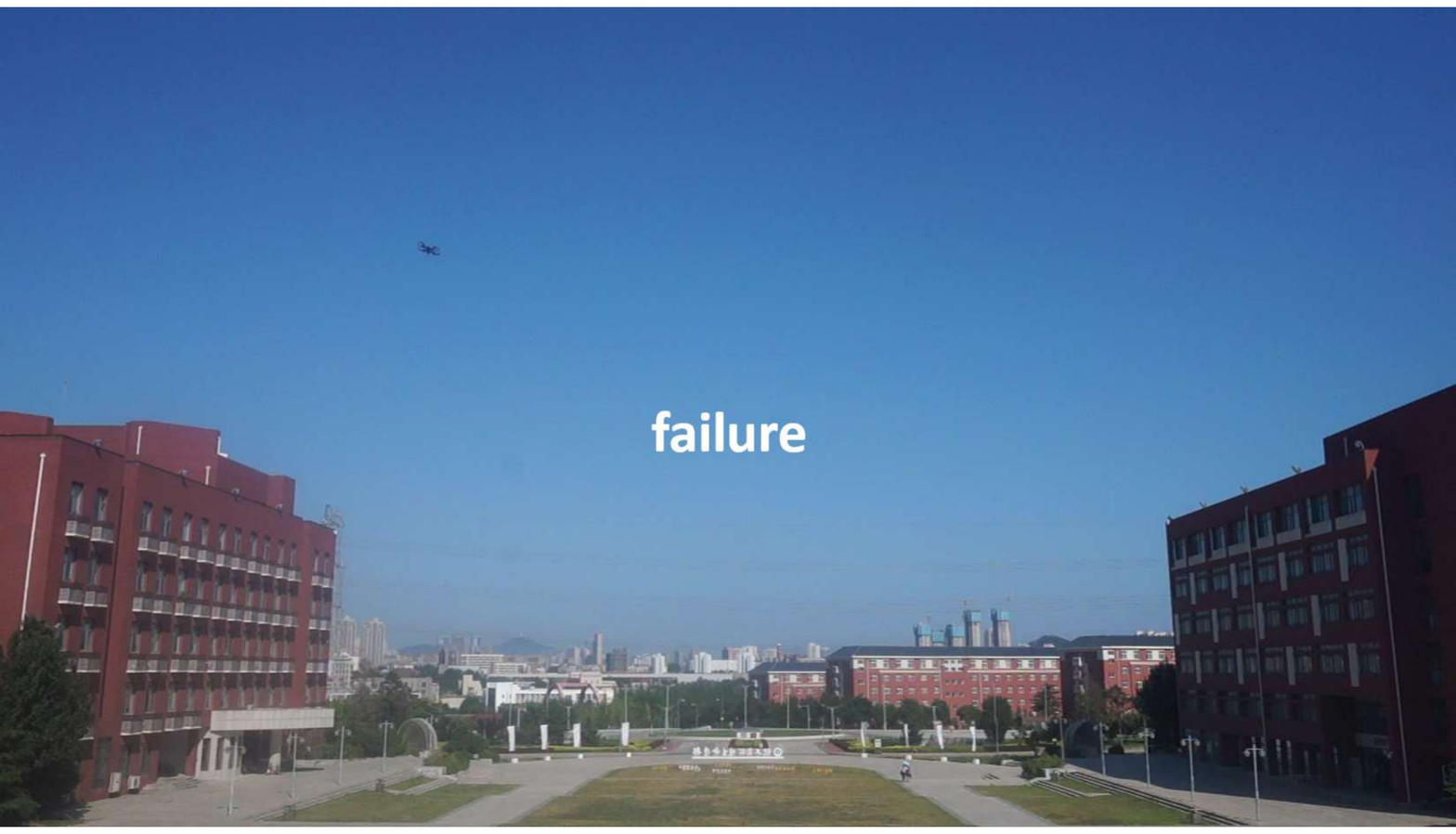}
    \end{subfigure}

    \begin{subfigure}[t]{0.12\textwidth}
        \includegraphics[angle=270,width=\linewidth]{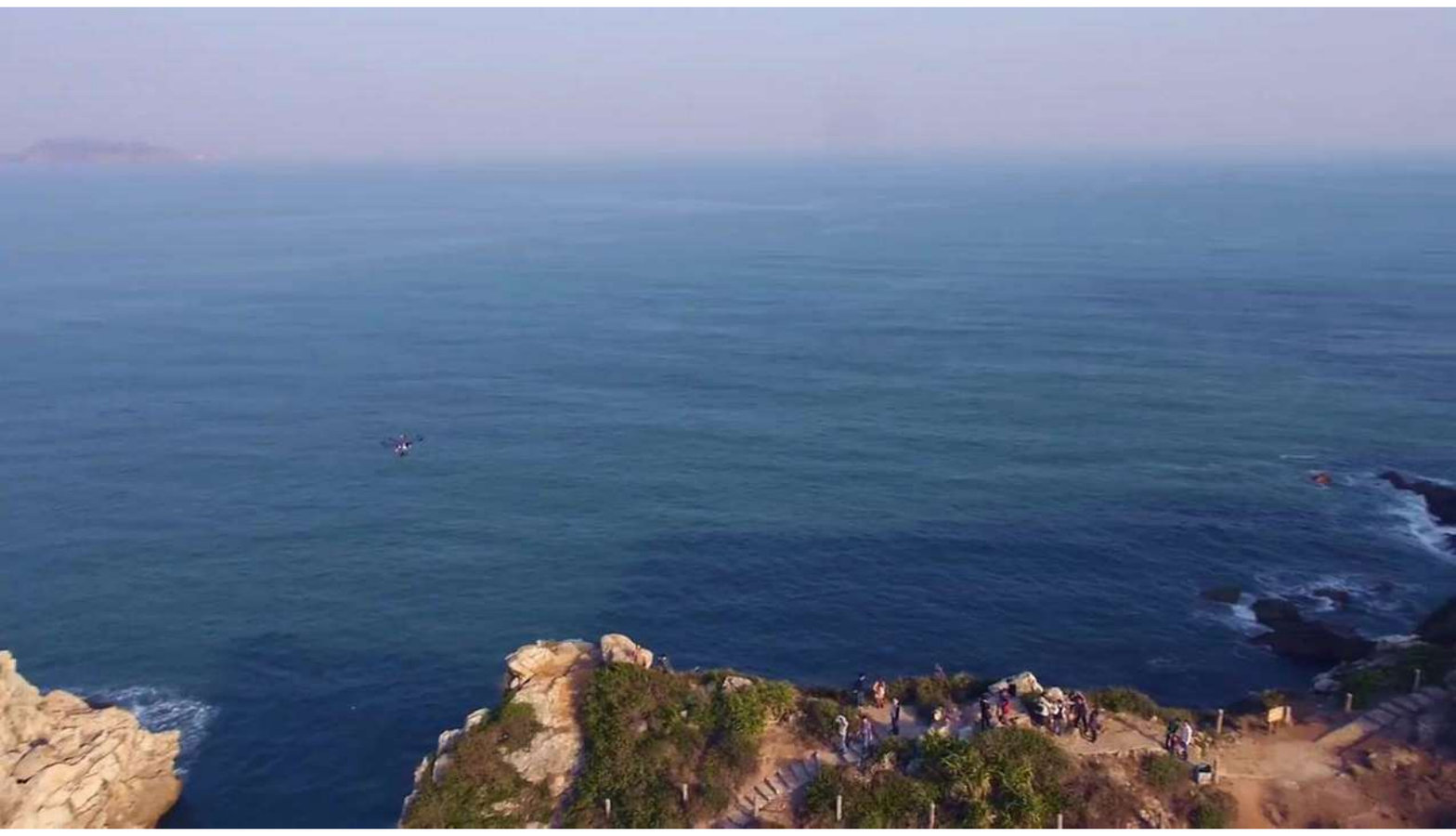}
        \caption{\tiny{Original}}
    \end{subfigure}
    \begin{subfigure}[t]{0.12\textwidth}
        \includegraphics[angle=270,width=\linewidth]{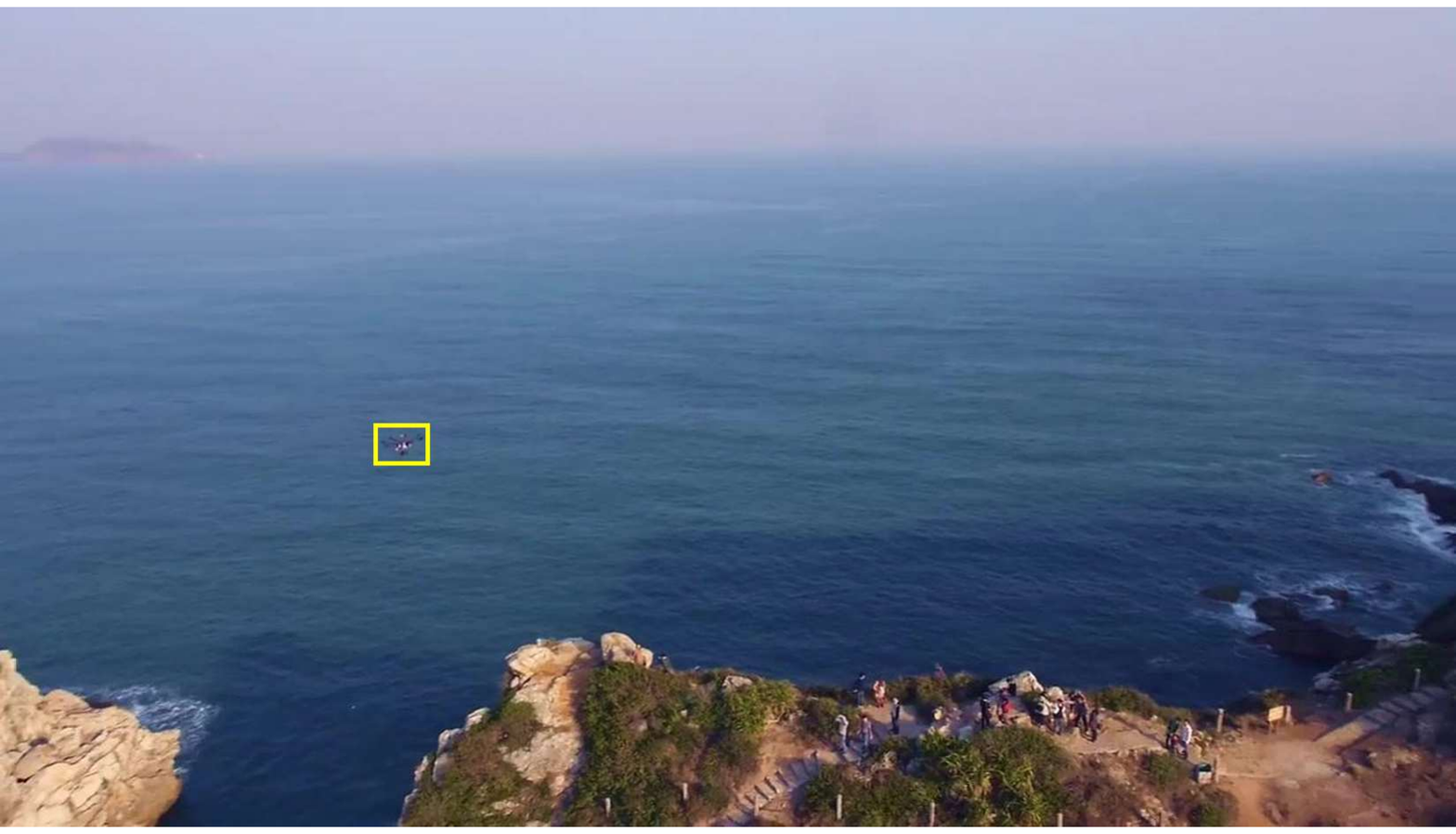}
        \caption{\tiny{Yolov12s-CBi}}
    \end{subfigure}
    \begin{subfigure}[t]{0.12\textwidth}
        \includegraphics[angle=270,width=\linewidth]{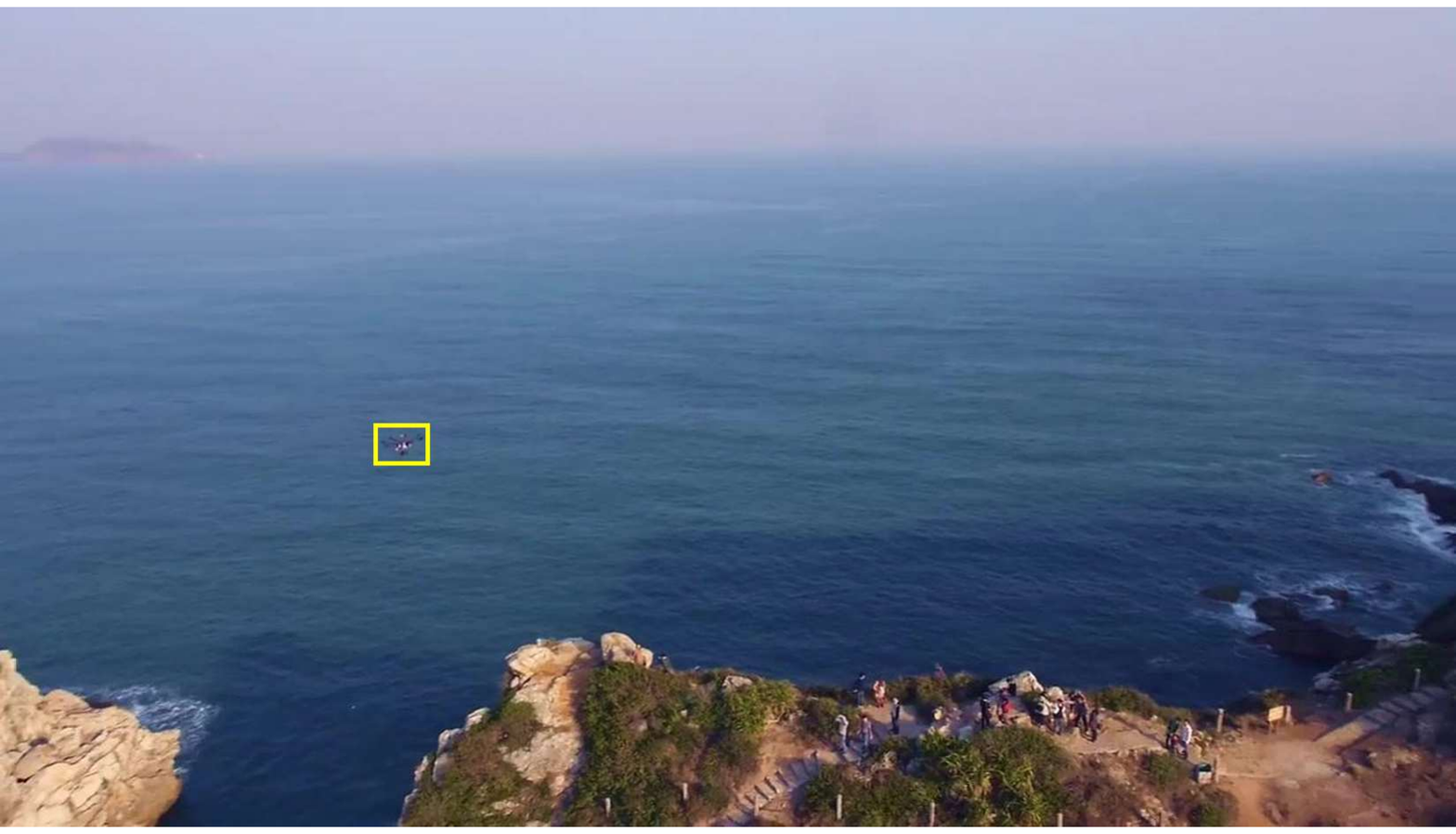}
        \caption{\tiny{Yolov8s-CBi}}
    \end{subfigure}
    \begin{subfigure}[t]{0.12\textwidth}
         \includegraphics[angle=270,width=\linewidth]{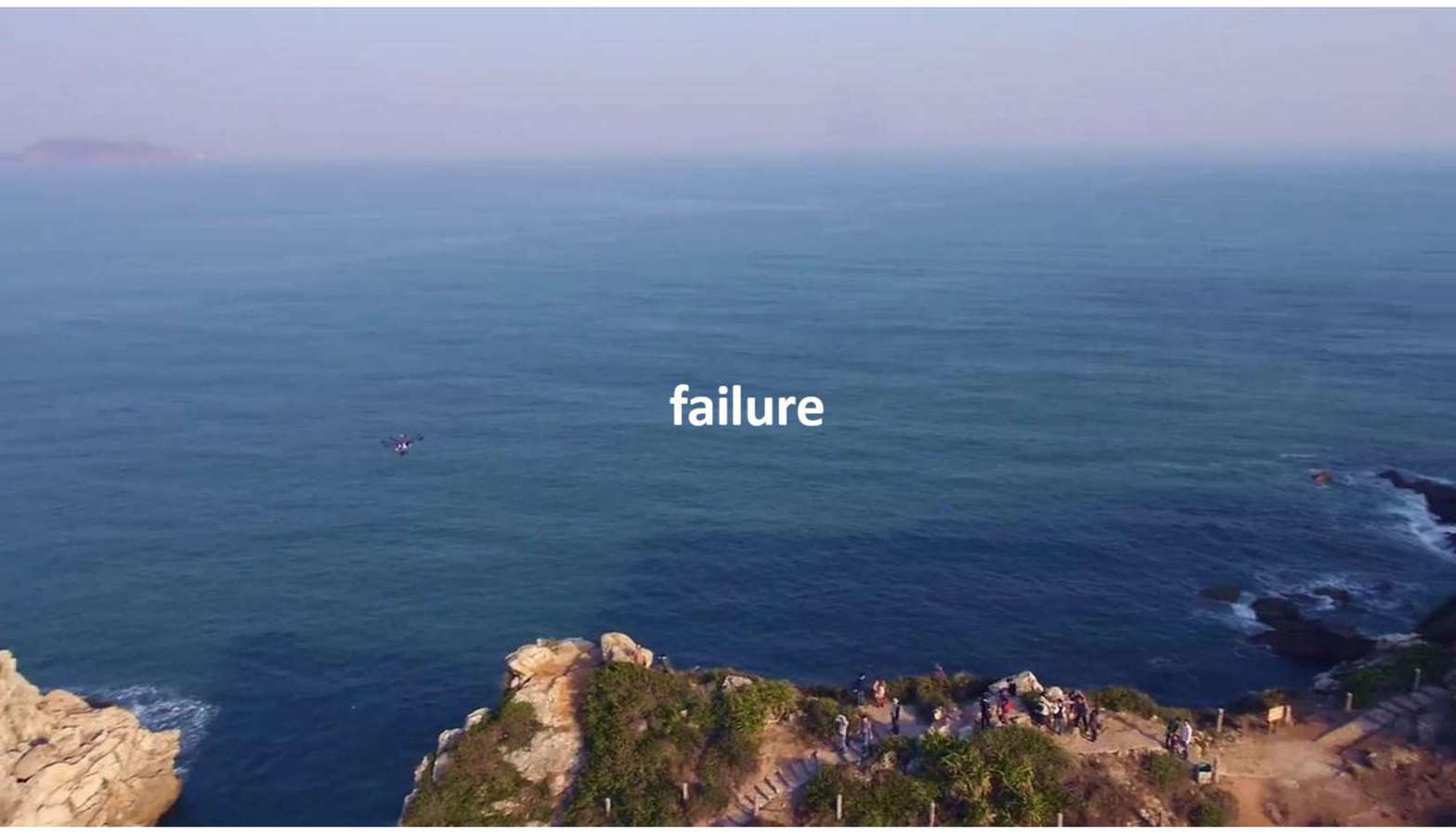}
        \caption{\tiny{Yolov9s-CBi}}
    \end{subfigure}
    \begin{subfigure}[t]{0.12\textwidth}
        \includegraphics[angle=270,width=\linewidth]{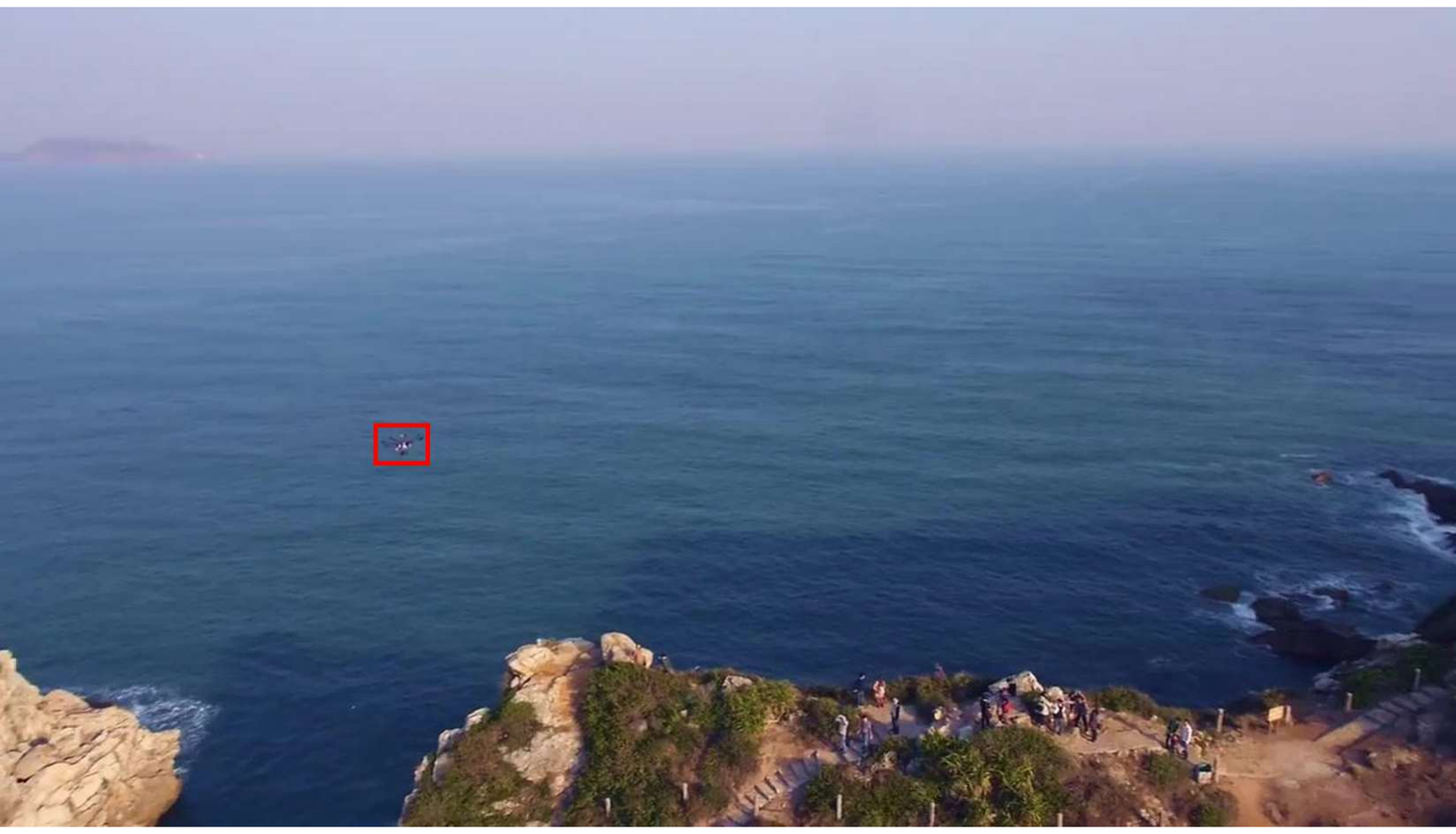}
        \caption{\tiny{Yolov10s-CBi}}
    \end{subfigure}
    \begin{subfigure}[t]{0.12\textwidth}
        \includegraphics[angle=270,width=\linewidth]{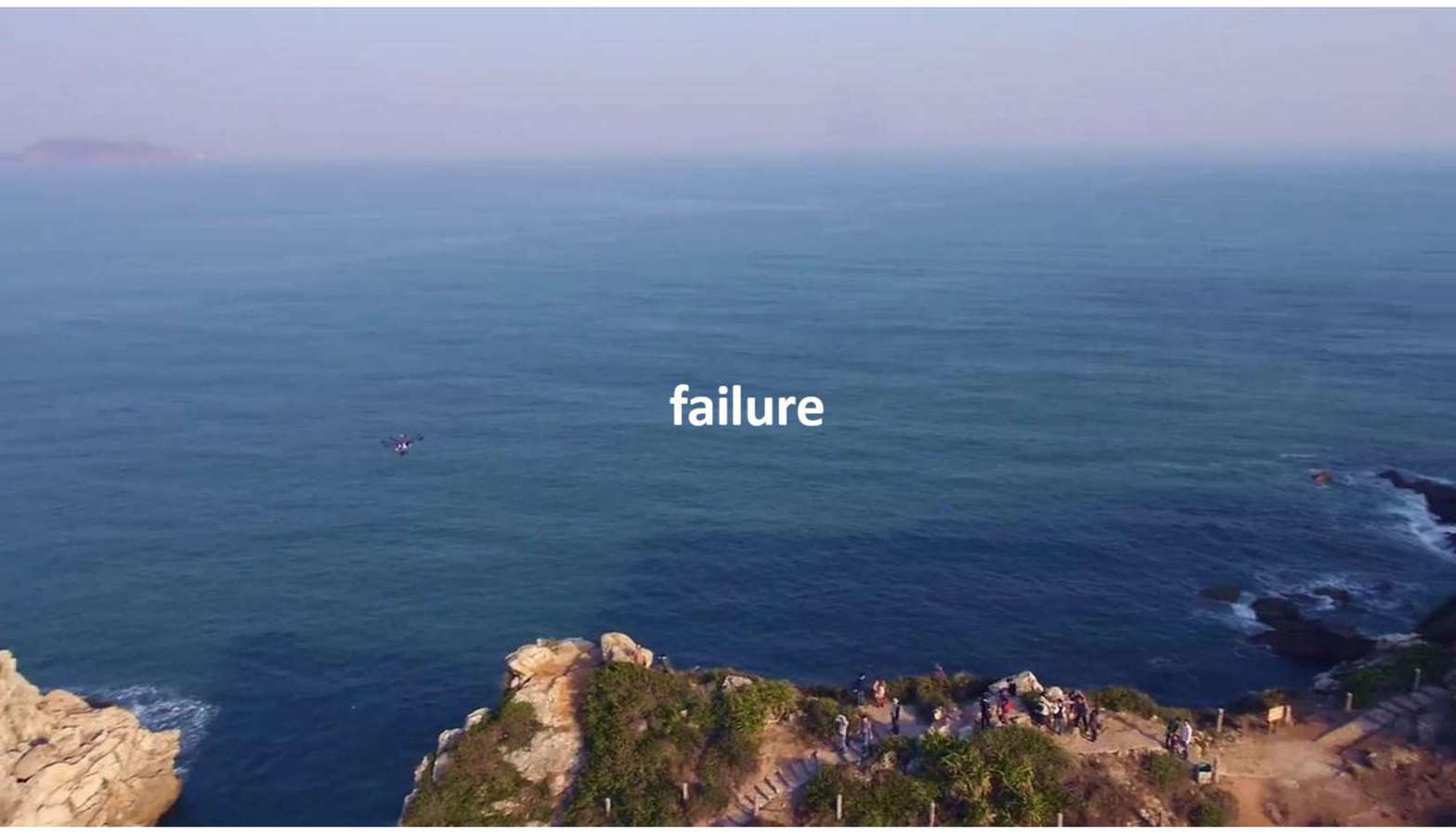}
        \caption{\tiny{Yolov11s-CBi}}
    \end{subfigure}
    \begin{subfigure}[t]{0.12\textwidth}
        \includegraphics[angle=270,width=\linewidth]{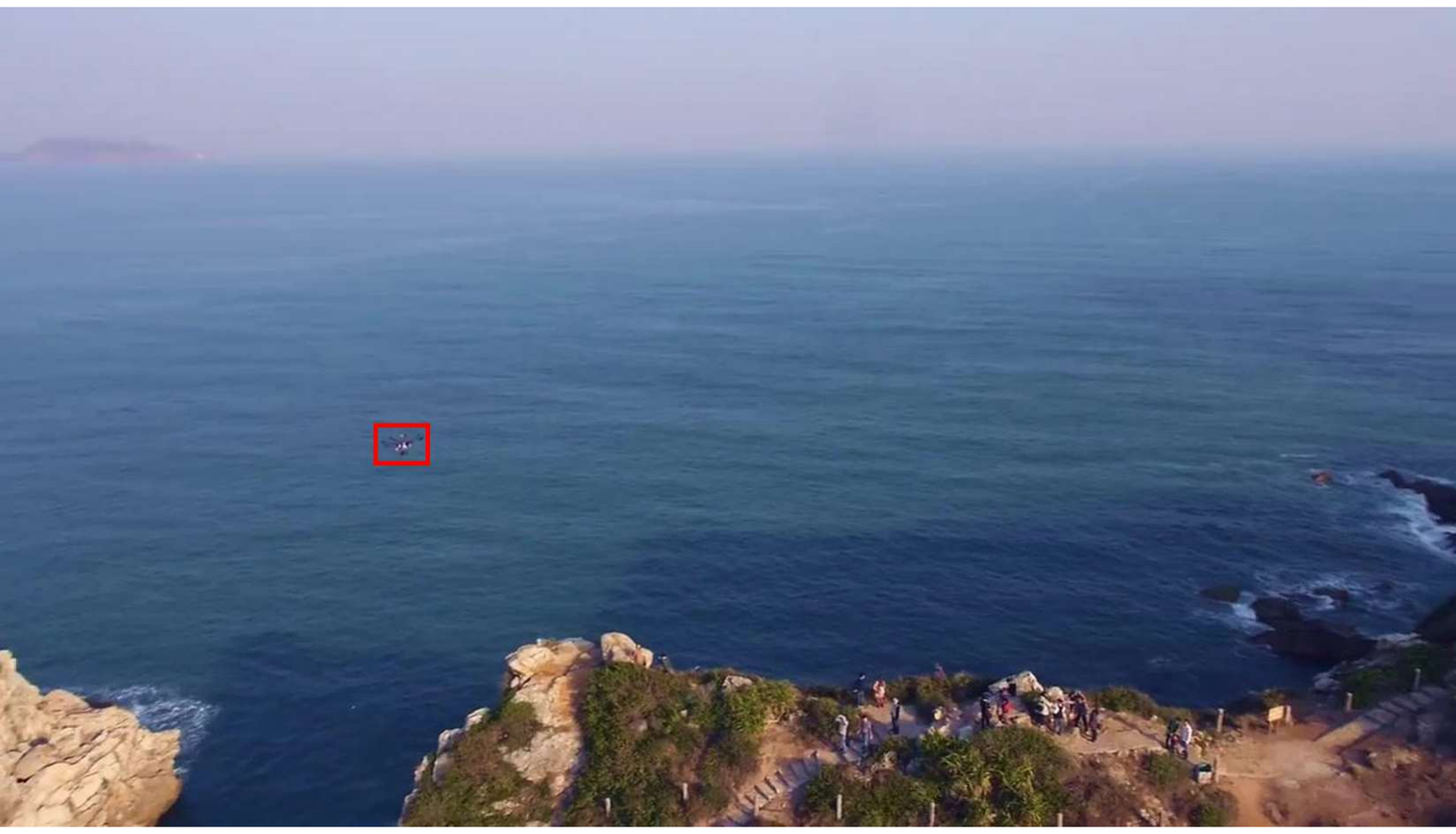}
        \caption{\tiny{Yolov12s-CBi}}
    \end{subfigure}
    \caption{Inference on the DUT-AntiUAV dataset using various YOLOvN-CBi variants. Yellow bounding boxes indicate true positives. Red bounding boxes represent false cases where a drone is misclassified as a bird. Purple bounding boxes indicate false cases where a background object is incorrectly identified as a drone. Failure cases where model fails to detect any objects are highlighted in white text.}
    \label{fig:inf_dut}
\end{figure*}

\section{Performance Analysis}\label{sec:experiments}

Although recent Yolo versions (Yolov5 through Yolov12) incorporate advanced features and improved training techniques, the assumption that newer versions always perform better for all applications is often misleading. In practice, the effectiveness of a model can vary significantly depending on the application. Importantly, the trade-off between accuracy and speed becomes critical for small drone detection in edge devices for real-time application.

In this study, the performance of the proposed YolovN-CBi architecture is compared with its base YolovN architecture, for N = 5, 8, 9, 10, 11, 12 which are selected based on their popularity and frequent use in recent work in drone detection and computer vision \cite{93,65,51}. The goal is to analyze the relative strengths and weaknesses of these models in the context of drone detection and to determine which model offers the best balance between detection accuracy and computational efficiency. All Yolo models and the corresponding CBi variants are trained on the Flying Object dataset introduced in Section \ref{sec:dataset}, using identical hyperparameters and data augmentations to ensure consistency. Training was performed on an NVIDIA A100 GPU with CUDA 11.6, using a batch size of 16 and an input image resolution of 1920×1080. The trained models are evaluated using five test datasets mentioned in Section \ref{sec:dataset}. . 

\subsection{Evaluation Metrics}
\label{subsec:evaluation_metric}

The performance of variants of the YOLO model is evaluated using standard object detection metrics: \textbf{Precision}, \textbf{Recall}, \textbf{mAP@0.5}, \textbf{mAP@0.5:0.95}, and \textbf{Inference time}.

\begin{itemize}
    \item \textbf{Precision} measures the proportion of correctly predicted positive detections among all positive predictions. It is defined as:
    \[
        \text{Precision} = \frac{TP}{TP + FP}
    \]
    where \( TP \) is the number of true positives and \( FP \) is the number of false positives. High precision indicates fewer false alarms.

    \item \textbf{Recall} quantifies the proportion of correctly predicted positive detections out of all actual positive instances:
    \[
        \text{Recall} = \frac{TP}{TP + FN}
    \]
    where \( FN \) is the number of false negatives. A high recall score indicates that the model does not miss the actual objects.

    \item \textbf{mAP@0.5} (mean Average Precision at IoU threshold 0.5) is a commonly used detection metric that averages precision across all object classes at a fixed Intersection over Union (IoU) threshold of 0.5.

    \item \textbf{mAP@0.5:0.95} computes the mean Average Precision across multiple IoU thresholds ranging from 0.5 to 0.95 with a step of 0.05. This metric provides a more rigorous evaluation considering detection quality across stricter localization thresholds.

    \item \textbf{Inference time (ms)} refers to the average time taken by the model to process a single frame, measured in milliseconds. This is critical for real-time deployment scenarios.
\end{itemize}

\begin{table*}[t]
\centering
\scriptsize 
\caption{Detailed performance metrics for various YOLO models and proposed models on the Flying Object Detection dataset.}
\label{tab:yolo_val}

\renewcommand{\arraystretch}{1.0}
\setlength{\tabcolsep}{4pt}

\begin{tabularx}{\textwidth}{l*{5}{>{\centering\arraybackslash}X}}
\toprule
\textbf{Model} & \textbf{mAP@0.5} & \textbf{mAP@0.5:0.95} & \textbf{Precision} & \textbf{Recall} & \textbf{Inference time (ms)} \\
\midrule
Yolov5s      & \textbf{0.993} & 0.687 & 0.988 & \textbf{0.986} & \textbf{3.9} \\
Yolov5s-CBi  & 0.973 & 0.616 & 0.943 & 0.942 & 6.5 \\
\midrule
Yolov8s      & 0.944 & 0.714 & 0.995 & 0.890 & 6.2 \\
Yolov8s-CBi  & 0.958 & \textbf{0.725} & 0.996 & 0.917 & 6.8 \\
\midrule
Yolov9s      & 0.899 & 0.687 & 0.997 & 0.799 & 8.3 \\
Yolov9s-CBi  & 0.927 & 0.703 & 0.996 & 0.857 & 9.9 \\
\midrule
Yolov10s     & 0.965 & 0.721 & 0.991 & 0.933 & 8.9 \\
Yolov10s-CBi & 0.949 & 0.711 & 0.990 & 0.901 & 10.1 \\
\midrule
Yolov11s     & 0.936 & 0.711 & 0.997 & 0.873 & 6.6 \\
Yolov11s-CBi & 0.938 & 0.712 & 0.996 & 0.878 & 8.7 \\
\midrule
Yolov12s     & 0.887 & 0.679 & \textbf{0.998} & 0.775 & 26.1 \\
Yolov12s-CBi & 0.884 & 0.680 & 0.997 & 0.770 & 24.9 \\
\bottomrule
\end{tabularx}
\end{table*}

\subsection{Impact of CBi on Training}\label{sec:cbi_vis}

Various small Yolo architectures identified as YolovNs are chosen to study the impact of the addition of the CBi module. These models are trained on the Flying Object Dataset and Table\ref{tab:yolo_val} shows the comparison of each YolovNs model and its CBi variant (YolovNs-CBi) on the Flying Object Validation set. It can be seen that Yolov5s (N = 5) model and its CBi variant outperform all other models in mAP@0.5 (0.993) and recall. They also have a comparable mAP@0.5:0.95 (0.687) with a low inference time of 3.9\,ms. Although newer models such as Yolov10s and Yolov12s show improvements in precision, their increased computational cost outweighs the benefits for real-time deployment scenarios.

\subsection{Cross-Dataset Analysis}

The models trained on Flying Object dataset are used to perform inference on the five test datasets. Table \ref{tab:yolo_inf} shows the comparison between the recall and the inference time on these five test datasets.  Yolov5s and its CBi variant achieve consistently higher recall values on most datasets, with only a moderate increase in inference time (to ~10.5\,ms). As evident, Yolov5s-CBi outperforms all the models by a huge margin on Local Test Data, which is the dataset for tiny objects. In contrast, newer models such as Yolov10s and Yolov12s, despite their architectural complexity, fail to generalize effectively, particularly on the Local Test Data with small drones, where recall drops drastically. 

The inference results of the evaluation are shown in Figure  \ref{fig:inf_dut} , which shows that Yolov5s-CBi detects drones that appear as small as 20 pixels and are missed or misclassified by other models. 

\begin{table*}
\centering
\caption{Inference Time and Recall for Yolo Models and Proposed Models on five test datasets}
\label{tab:yolo_inf}
\renewcommand{\arraystretch}{1.0} 
\footnotesize              
\setlength{\tabcolsep}{3pt} 
\begin{tabularx}{\textwidth}{l*{10}{>{\centering\arraybackslash}X}}
\toprule
\textbf{Model} & 
\multicolumn{2}{c}{\textbf{AntiUAV}} & 
\multicolumn{2}{c}{\textbf{ UAV-CDT}} & 
\multicolumn{2}{c}{\textbf{DUT-AntiUAV}} & 
\multicolumn{2}{c}{\textbf{Local Test Data}} & 
\multicolumn{2}{c}{\textbf{LRDD}} \\
\cmidrule(lr){2-3}
\cmidrule(lr){4-5}
\cmidrule(lr){6-7}
\cmidrule(lr){8-9}
\cmidrule(lr){10-11}
& Inf. time(ms) & Recall 
& Inf. time(ms) & Recall 
& Inf. time(ms) & Recall 
& Inf. time(ms) & Recall 
& Inf. time(ms) & Recall \\
\midrule
Yolov5s & 6.8 & \textbf{0.873} & 7.0 & \textbf{0.967} & 6.8 & \textbf{0.724} & 6.7 & 0.199 & 6.8 & \textbf{0.558} \\
Yolov5s-CBi & 10.5 & 0.870 & 10.9 & 0.926 & 10.5 & 0.293 & 10.5 & \textbf{0.559} & 10.6 & \textbf{0.589} \\
\midrule
Yolov8s & 7.8 & 0.447 & 8.0 & 0.312 & 7.9 & 0.448 & 7.8 & 0.006 & 7.8 & 0.338 \\
Yolov8s-CBi & 10.9 & 0.261 & 11.0 & 0.251 & 10.8 & 0.396 & 10.9 & 0.003 & 10.9 & 0.314 \\
\midrule
Yolov9s & 20.9 & 0.495 & 21.1 & 0.336 & 21.1 & 0.469 & 21.1 & 0.129 & 21.2 & 0.414 \\
Yolov9s-CBi & 21.9 & 0.500 & 21.8 & 0.318 & 21.7 & 0.465 & 22.1 & 0.357 & 21.7 & 0.366 \\
\midrule
Yolov10s & 10.9 & 0.260 & 11.1 & 0.287 & 11.1 & 0.425 & 11.0 & 0.027 & 11.0 & 0.304 \\
Yolov10s-CBi & 12.0 & 0.405 & 12.3 & 0.305 & 12.2 & 0.427 & 12.2 & 0.005 & 12.0 & 0.303 \\
\midrule
Yolov11s & 9.6 & 0.255 & 10.0 & 0.289 & 9.9 & 0.455 & 9.9 & 0.006 & 9.8 & 0.334 \\
Yolov11s-CBi & 10.2 & 0.468 & 10.6 & 0.288 & 10.5 & 0.435 & 10.4 & 0.264 & 10.4 & 0.336 \\
\midrule
Yolov12s & 13.7 & 0.475 & 13.6 & 0.406 & 14.0 & 0.471 & 13.9 & 0.005 & 13.7 & 0.405 \\
Yolov12s-CBi & 14.2 & 0.473 & 14.4 & 0.378 & 14.4 & 0.466 & 14.4 & 0.008 & 14.4 & 0.423 \\
\bottomrule
\end{tabularx}
\end{table*}

\subsection{Comparison of CBi Variants}
The five proposed CBi variants of the YOLOv5s architecture are evaluated on popular DUT-AntiUAV benchmark dataset. Table~\ref{tab:cbi_variants} summarizes the configuration and performance of each variant in terms of model complexity (measured by parameter count) and detection accuracy (measured by mAP@0.5).

\begin{figure*}
    \centering
    \includegraphics[width=0.9\textwidth]{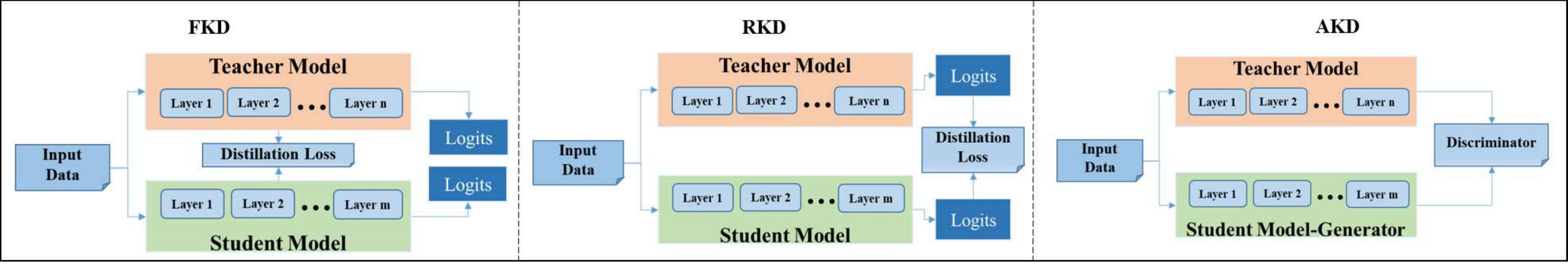}
    \caption{Distillation Strategies} 
    \label{fig:akd_diagram}
\end{figure*}

\begin{table}[t]
\centering
\caption{Comparison of Yolov5s-CBi Variants on DUT-AntiUAV Dataset}
\label{tab:cbi_variants}
\begin{tabular}{lcc}
\toprule
\textbf{Variant} & \textbf{Params (M)} & \textbf{mAP@0.5} \\
\midrule
Yolov5s-CBi & 17.69 & 0.9143 \\
Yolov5s-CBi-v2 & 17.76 & \textbf{0.9545} \\
Yolov5s-CBi-v3 & 17.73 & 0.9523 \\
Yolov5s-CBi-v4 & 18.04 & 0.9444 \\
Yolov5s-CBi-v5 & \textbf{16.41} & 0.9287 \\
\bottomrule
\end{tabular}
\end{table}

\paragraph*{Analysis.}
Among the five variants, \textbf{CBi-v2} achieved the highest mAP@0.5 of \textbf{0.9545}, indicating that integrating CBAM early in the fusion path effectively enhances discriminative feature learning. Variant \textbf{CBi-v3} also performed well, suggesting that replacing key backbone modules with CBAM (C3b) provides benefits by improving low-level feature encoding.

Interestingly, although \textbf{CBi-v5} has the lowest parameter count (16.41M), its performance is better than Yolov5s-CBi but not better than others, pointing to a possible trade-off between aggressive module replacement and feature generalization. \textbf{CBi-v4}, which places CBAM after the BiFPN fusion stage, performs worse than earlier fusion strategies, suggesting that late refinement is less effective for small object detection.

Overall, these results highlight that the location of attention mechanisms significantly affects the detection accuracy, with early integration in the feature fusion process yielding the best results.

\section{Knowledge Distillation }
\label{sec:kd}

The proposed Yolov5-CBi variant is robust and accurate in small drone detection task; however, its computational demands (twice that of the base model) make it less ideal for real-time inference on resource-constrained platforms or embedded edge systems. This is mitigated by using a technique called Knowledge Distillation (KD), which transfers knowledge from a high-capacity teacher model to a compact student model, retaining performance while reducing the inference time.

Knowledge distillation is performed between YOLOv5m (teacher model) and YOLOv5n (student model). This choice is motivated by the architectural and computational gap between the two, which is critical for effective knowledge transfer. Although YOLOv5s appears to be a closer match to YOLOv5n in terms of scale, the differences in model capacity are relatively minor, both use a depth multiplier of 0.33, and width multipliers are 0.50 for Yolov5s and 0.25 for Yolov5n, respectively. This minor gap may constrain the potential benefits of distillation, as the teacher may not offer sufficiently richer feature representations. In contrast, YOLOv5m employs a deeper and wider architecture (depth multiplier: 0.67, width multiplier: 0.75), providing a more expressive and informative supervisory signal. This larger capacity allows the teacher model to encode more nuanced spatial and semantic information, which can then be effectively distilled into the compact YOLOv5n student model. Thus, m-to-n pairing is considered the best for distillation because of the diversity of the models and the feasibility for lightweight deployment.

Two teacher-student configurations were explored:
\begin{enumerate}
    \item \textbf{Student-only CBi variant}: A regular Yolov5m model is teacher and is distilled into a student model, which is enhanced with CBi i.e., Yolov5n-CBi. This configuration is termed YolovNn-CBi-Distill-Yolov5Nm.
    \item \textbf{Dual-CBi configuration}: A CBi variant Yolov5m-CBi model is selected as a teacher and is distilled into a student model, which is also enhanced with CBi i.e., Yolov5n-CBi. This configuration is termed YolovNn-CBi-Distill-Yolov5Nm-CBi.    
\end{enumerate}

The distillation of these models is achieved using three strategies:
\begin{enumerate}
    \item \textbf{Feature-based Knowledge Distillation (FKD)}: 
    This strategy focuses on transferring deep feature representations between teacher and student models. The intermediate features of the teacher model are used to guide the learning process of the student model. FKD preserves the structural similarity between the feature maps of both models by minimizing the loss between intermediate features of the teacher and student at multiple layers as shown in Figure~\ref{fig:akd_diagram}.
    
    \item \textbf{Response-based Knowledge Distillation (RKD)}: 
    RKD focuses on matching the output logits of the teacher and student models by guiding the student network to mimic the teacher's behavior. The distillation loss in RKD is computed by comparing the logits between the teacher and student models, as depicted in Figure \ref{fig:akd_diagram}. 
    
    \item \textbf{Adversarial Knowledge Distillation (AKD)}: 
    AKD introduces an additional component to the distillation process: a discriminator. This strategy treats the teacher-student feature map relationship as an adversarial game. The discriminator differentiates between the feature maps of teacher and student. The objective of the student model is to fool the discriminator by generating feature maps that are indistinguishable from those of the teacher. In AKD, the distillation loss consists of the discriminator loss between the real (teacher) and fake (student) features, as shown in Figure \ref{fig:akd_diagram}. 
\end{enumerate}

\begin{table*}

\scriptsize                          
\caption{Performance comparison of student models trained with different KD strategies.}
\label{tab:kd_val}
\renewcommand{\arraystretch}{1.1}
\setlength{\tabcolsep}{3pt}          

\begin{threeparttable}
\begin{tabularx}{\textwidth}{
  >{\raggedright\arraybackslash}p{2cm}   
  >{\raggedright\arraybackslash}p{2.5cm}   
  *{5}{>{\centering\arraybackslash}X}      
}
\toprule
\textbf{Distillation strategy} & \textbf{Model} &
\textbf{mAP@0.5} & \textbf{mAP@0.5:0.95} &
\textbf{Precision} & \textbf{Recall} &
\textbf{Inf.\ time (ms)} \\
\midrule
Base Models &
Yolov5n         & 0.9810 & 0.6370 & 0.9730 & 0.9760 & 2.0 \\
& Yolov5m       & 0.9741 & 0.6171 & 0.9447 & 0.9443 & 14.0 \\
\midrule
FKD &
Yolov5n-CBi-Distill-Yolov5m             & 0.9761 & 0.6084 & 0.9603 & 0.9599 & 3.2 \\
& Yolov5n-CBi-Distill-Yolov5m-CBi       & 0.9674 & 0.5994 & 0.9514 & 0.9465 & 2.6 \\
\midrule
RKD &
Yolov5n-CBi-Distill-Yolov5m             & 0.9803 & 0.6225 & 0.9723 & 0.9735 & 3.0 \\
& Yolov5n-CBi-Distill-Yolov5m-CBi       & 0.9743 & 0.6159 & 0.9475 & 0.9524 & 3.3 \\
\midrule
AKD &
Yolov5n-CBi-Distill-Yolov5m             & 0.9678 & 0.6028 & 0.9445 & 0.9424 & 3.2 \\
& Yolov5n-CBi-Distill-Yolov5m-CBi       & \textbf{0.9902} & \textbf{0.6573} &
                                           \textbf{0.9835} & \textbf{0.9842} & 2.4 \\
\bottomrule
\end{tabularx}

\vspace{1mm}
\begin{tablenotes}[flushleft]
\footnotesize
\item FKD: Feature-based Knowledge Distillation
 \item RKD: Response-based Knowledge Distillation
 \item AKD: Adversarial Knowledge Distillation.
\item Yolov5n-CBi-Distill-Yolov5m: student distilled from standard Yolov5m teacher.
\item Yolov5n-CBi-Distill-Yolov5m-CBi: student distilled from CBi variant of Yolov5m teacher.
\end{tablenotes}
\end{threeparttable}
\end{table*}

\subsection{Evaluation of Distilled Models}
\label{sec:KDresults}

This section analyzes the performance of distilled Yolov5n-CBi models from two teachers and following three KD strategies, FKD, RKD and AKD, across five test datasets.

The distillation is performed following the above mentioned three strategies using Flying Object dataset. Table \ref{tab:kd_val} shows the comparison of the performance metrics between the base models and the distilled models. It can be seen that the base model, Yolov5n, achieves a mAP@0.5 of 0.981 with an inference time of 2\,ms. Among all distilled models, Yolov5n-CBi distilled from the Yolov5m-CBi teacher model using the AKD strategy achieves the highest mAP@0.5 of 0.9902 and mAP@0.5:0.95 of 0.6573, with the lowest inference time of 2.4\,ms.

\begin{table*}
\caption{Inference time and recall for different KD strategies on five test datasets}
\label{tab:kd_inf}

\tiny
\setlength{\tabcolsep}{2pt} 
\renewcommand{\arraystretch}{1.0}

\begin{tabularx}{\textwidth}{ll*{10}{>{\centering\arraybackslash}X}}
\toprule
\textbf{Distillation Strategy} & \textbf{Model} &
\multicolumn{2}{c}{\textbf{AntiUAV}} &
\multicolumn{2}{c}{\textbf{UAV-CDT}} &
\multicolumn{2}{c}{\textbf{DUT-AntiUAV}} &
\multicolumn{2}{c}{\textbf{Local Test}} &
\multicolumn{2}{c}{\textbf{LRDD}} \\
\cmidrule(lr){3-4}
\cmidrule(lr){5-6}
\cmidrule(lr){7-8}
\cmidrule(lr){9-10}
\cmidrule(lr){11-12}
& & Inf. time & Recall
  & Inf. time & Recall
  & Inf. time & Recall
  & Inf. time & Recall
  & Inf. time & Recall \\
\midrule
FKD & Yolov5n-CBi-Distill-Yolov5m         & 10.3 & \textbf{0.962} & 10.7 & 0.840 & 10.3 & 0.554 & 10.2 & 0.090 & 10.2 & 0.560 \\
     & Yolov5n-CBi-Distill-Yolov5m-CBi    &  9.9 & 0.878          & 10.2 & 0.876 &  9.9 & 0.572 &  9.9 & 0.264 & 10.0 & 0.605 \\
RKD & Yolov5n-CBi-Distill-Yolov5m         & 10.3 & 0.950          & 10.7 & \textbf{0.904} & 10.3 & \textbf{0.578} & 10.2 & 0.741 & 10.2 & \textbf{0.612} \\
     & Yolov5n-CBi-Distill-Yolov5m-CBi    &  9.4 & 0.898          &  9.6 & 0.887 &  9.6 & 0.548 &  9.4 & 0.307 &  9.4 & 0.558 \\
AKD & Yolov5n-CBi-Distill-Yolov5m         & 10.2 & 0.920          & 10.8 & 0.880 & 10.4 & 0.567 & 10.1 & 0.798 & 10.3 & 0.576 \\
     & Yolov5n-CBi-Distill-Yolov5m-CBi    &  9.3 & 0.950          &  9.6 & 0.863 &  9.4 & 0.565 &  9.4 & \textbf{0.799} & 9.4 & 0.556 \\
\bottomrule
\end{tabularx}
\begin{tablenotes}
    \item Inf. time is Inference time measured in milliseconds
\end{tablenotes}
\end{table*}

It can be seen that in both the FKD and RKD strategies, the performance of the student model approaches that of the base teacher model.

\begin{figure*}
    \centering
    \begin{subfigure}[t]{0.16\textwidth}
        \includegraphics[angle=270,width=\linewidth, trim=0pt 150pt 180pt 150pt, clip]{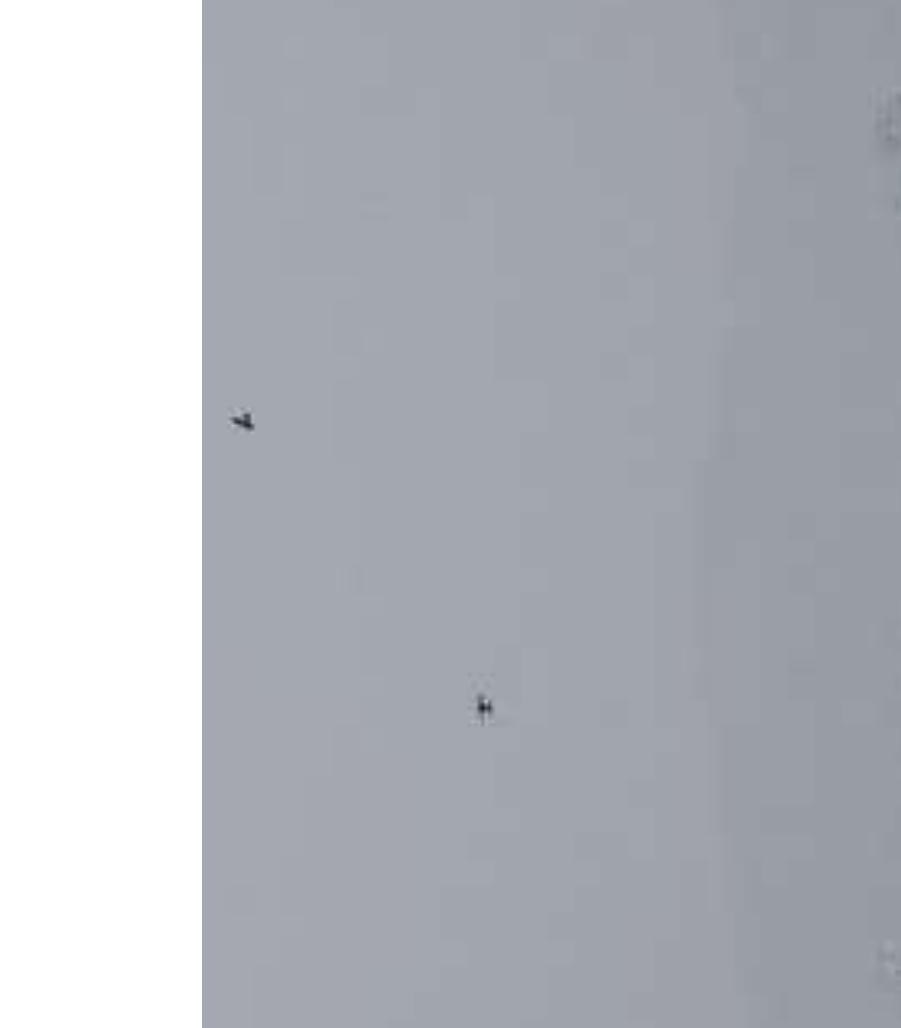}
        \caption{Original \\ Frame}
    \end{subfigure} 
    \begin{subfigure}[t]{0.16\textwidth}
        \includegraphics[angle=270,width=\linewidth, trim=0pt 150pt 180pt 150pt, clip]{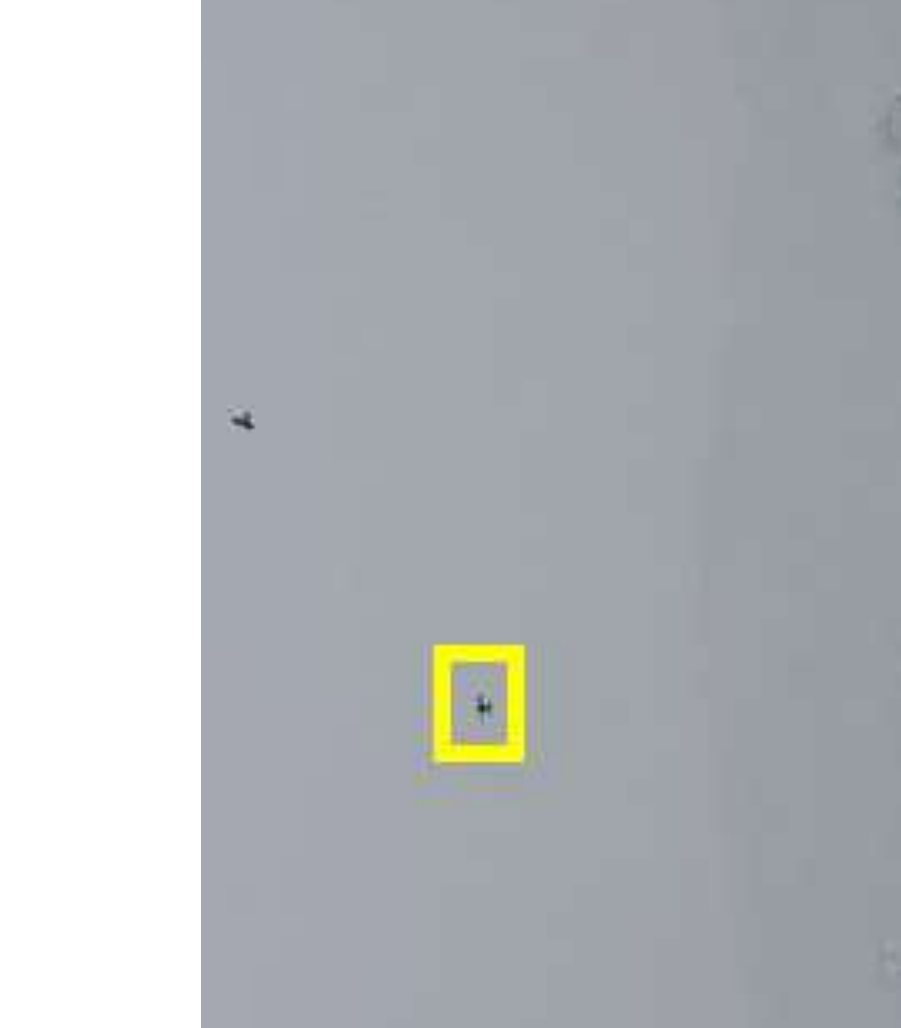}
        \caption{Teacher Model \\(Yolov5m-CBi)}
    \end{subfigure}
    \begin{subfigure}[t]{0.16\textwidth}

        \includegraphics[angle=270,width=\linewidth, trim=0pt 150pt 180pt 150pt, clip]{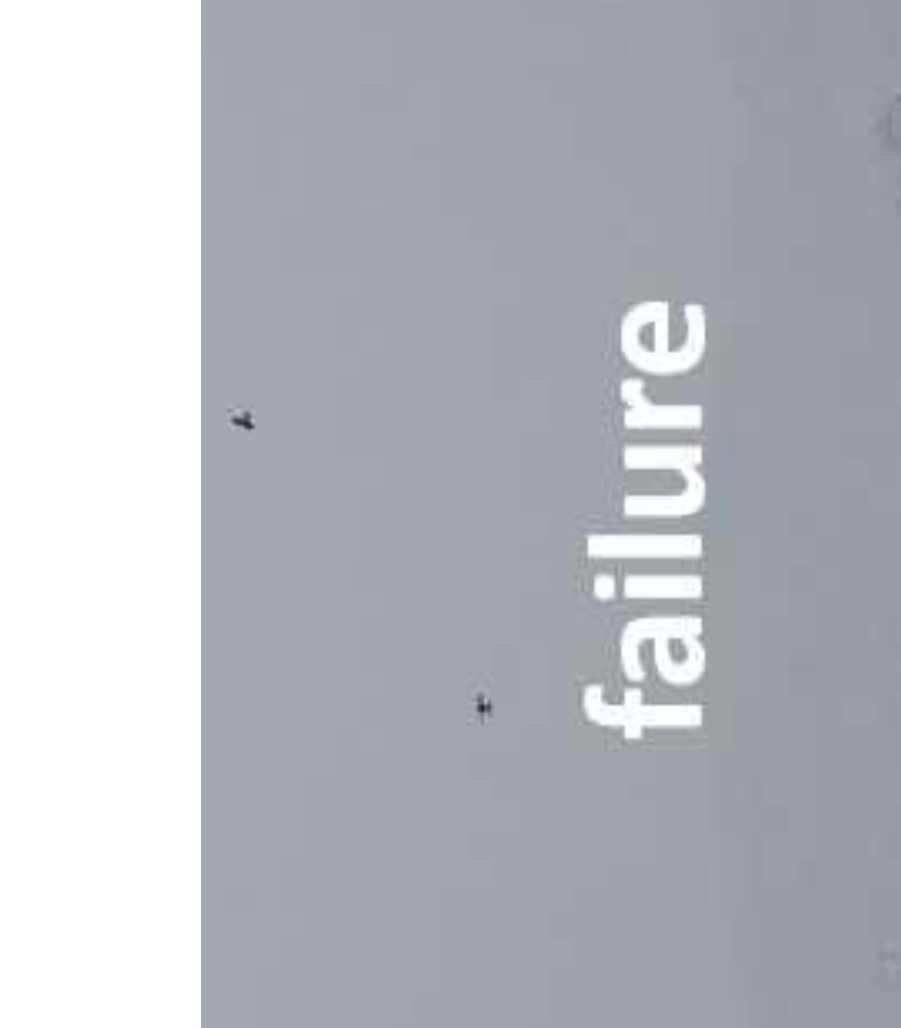}
        \caption{Base Model \\(Yolov5n-CBi)}
    \end{subfigure}
    \begin{subfigure}[t]{0.16\textwidth}
        \includegraphics[angle=270,width=\linewidth, trim=0pt 150pt 180pt 150pt, clip]{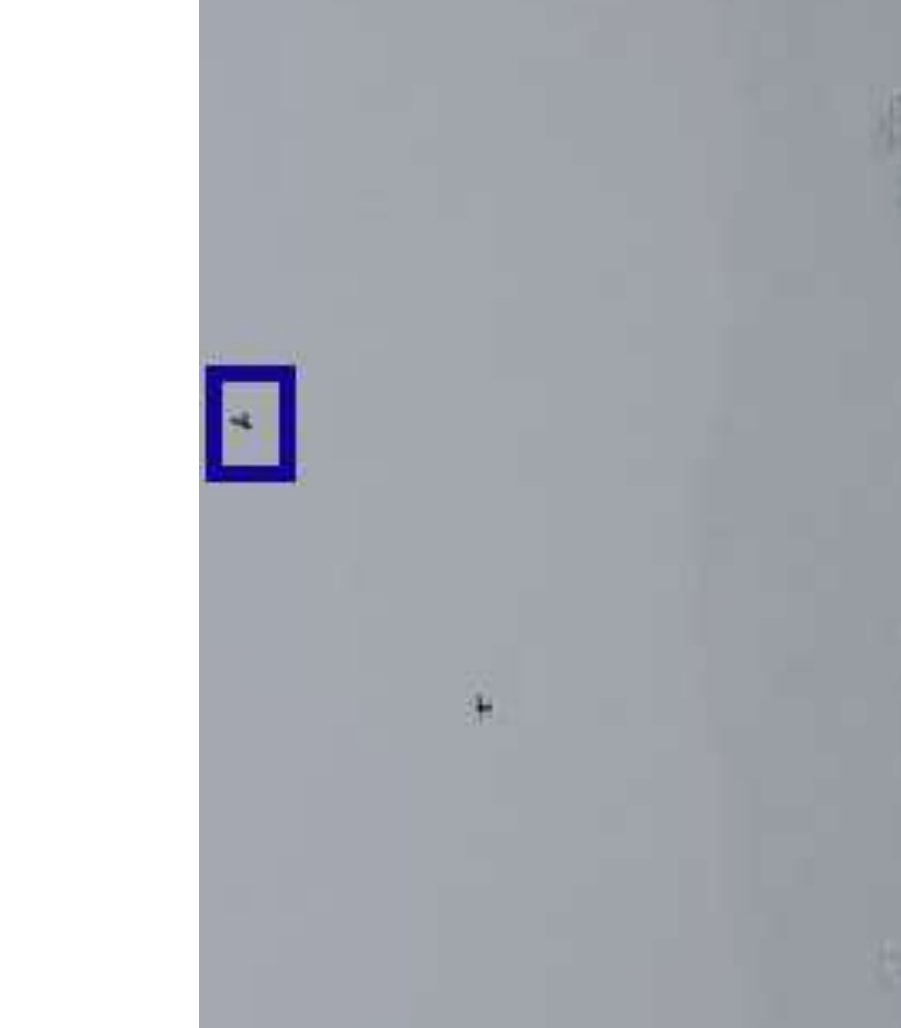}
        \caption{RKD Distilled model  \\(Yolov5n-CBi)}
    \end{subfigure}
    \begin{subfigure}[t]{0.16\textwidth}
        \includegraphics[angle=270,width=\linewidth, trim=0pt 150pt 180pt 150pt, clip]{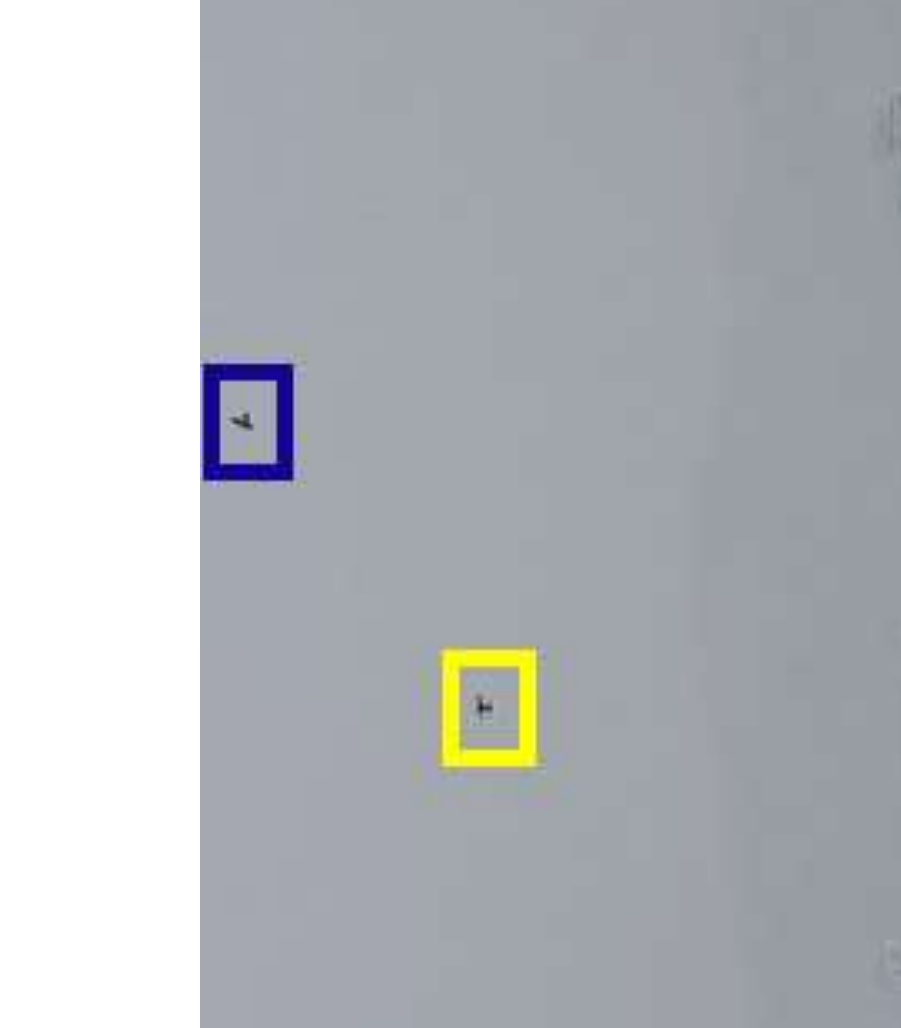}
        \caption{FKD Distilled model \\(Yolov5n-CBi)}
    \end{subfigure}
    \begin{subfigure}[t]{0.16\textwidth}
        \includegraphics[angle=270,width=\linewidth, trim=0pt 150pt 180pt 150pt, clip]{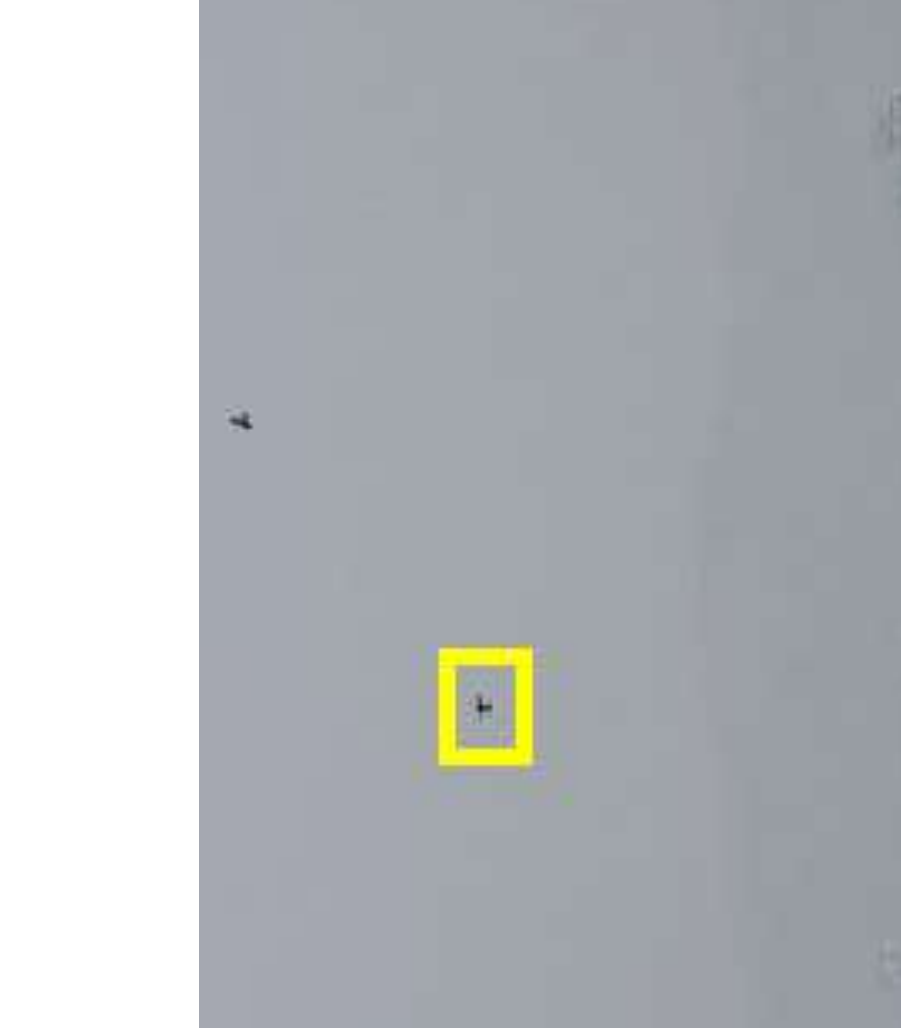}
        \caption{AKD Distilled model \\(Yolov5n-CBi)}
    \end{subfigure}
     
    \caption{Inference on Local test Dataset for different KD strategies. Yellow bounding boxes indicate true positives. Red bounding boxes represent false positives where a drone is misclassified as a bird. Purple bounding boxes indicate false positives where a background object is incorrectly identified as a drone. Failure cases where model fails to detect any objects are highlighted in red text.}
    \label{fig:inf_kd}
\end{figure*}

Table \ref{tab:kd_inf} shows the recall and inference time on five test datasets: AntiUAV,  UAV-CDT, DUT-AntiUAV, LRDD, and Local Test Data. Figure \ref{fig:inf_kd} shows a visual detection example from the local test data for all strategies. The FKD and RKD models fail to predict when the drone is very small, but AKD matches the output with the teacher model shown in Figure \ref{fig:inf_kd}. 

The AKD approach is compared with a state-of-the-art model trained on the UAV-CDT dataset \cite{53}, which reports an mAP of 0.990 and an inference time of 13\,ms. The distilled AKD model using a YOLOv5n backbone with CBi surpasses this benchmark by achieving an mAP of 0.9922 while reducing the inference time to 3.6\,ms. This not only demonstrates competitive accuracy, but also substantial improvements in speed, making the model highly suitable for latency-sensitive drone surveillance scenarios.

Student-Only-CBi variants generally achieve higher recall, particularly on datasets with very small objects like the Local Test Data (e.g., 0.741 recall using FKD). Among the knowledge distillation methods, Adversarial KD (AKD) offers the best balance between recall and speed. It can be seen that Yolov5n-CBi-Distill-Yolov5m-CBi reaches 0.950 recall on AntiUAV and 0.799 on Local Test Data, due to better feature alignment between teacher and student networks. AKD outperforms FKD and RKD when the teacher and student share similar architectures. Student-only CBi setups provide strong generalization and high recall at minimal cost, with distilled Yolov5n6-CBi models via AKD, matching the accuracy of the teacher, while also being faster. This makes them highly suitable for edge drone surveillance in real time, where both speed and sensitivity to small targets are critical.

\section{Comparison}
\label{sec:comparison}

The performance metrics of the proposed model are evaluated on the DUT-AntiUAV benchmark dataset \cite{20} and compared with several state-of-the-art drone detection models based on modified YOLO architectures.

\begin{table*}

\renewcommand{\arraystretch}{1.2}
\centering
\caption{Performance comparison of proposed model with state-of-the-art drone detection models on the DUT-AntiUAV \cite{20} dataset}
\label{tab:val_dut}
\renewcommand{\arraystretch}{1.0} 
\setlength{\tabcolsep}{4pt} 
\begin{tabularx}{\textwidth}{l*{10}{>{\centering\arraybackslash}X}}
\toprule
\textbf{Model} & \textbf{mAP@0.5} & \textbf{mAP@0.5:0.95} & \textbf{Precision} & \textbf{Recall} \\
\midrule
YOLOv5(baseline) & 0.9231	&0.6274	&0.9641	&0.8504
 \\
LA-YOLO\cite{97} & 0.929 & - & 0.944 & 0.867 \\
SEB-YOLOv8s\cite{99} & 0.905 & 0.615 & 0.959 & 0.831 \\
DRBD-YOLOv8s\cite{98} & 0.951 & 0.554 & 0.946 & 0.920 \\
IASL-YOLO\cite{95} & 0.924 & 0.619 & 0.96 & 0.882 \\
\hdashline
\textbf{Proposed Models}\\
YOLOv5s-CBi & 0.9143 & 0.5981 &	0.9491 & 0.8512\\
YOLOv5s-CBi-v2 & \textbf{0.9545} &	\textbf{0.6711} &	\textbf{0.9716} & \textbf{0.9262}\\
YOLOv5s-CBi-v3 & 0.9523 &	0.6577 &	0.958 & 0.9225\\
YOLOv5s-CBi-v4 & 0.9441 & 0.6427 & 0.9604 & 0.9103 \\
YOLOv5s-CBi-v5 & 0.9287 & 0.6253 & 0.9588 & 0.8709 \\
\hdashline
\textbf{AKD Distilled Models}\\
Yolov5n-CBi-Distill-Yolov5m-CBi & 0.9357 &	0.6417	& 0.963	&0.8935 \\
Yolov5n-CBi-v2-Distill-Yolov5m-CBi & 0.9323	& 0.6356	&0.9685	&0.8798	 \\
Yolov5n-CBi-v3-Distill-Yolov5m-CBi & 0.9376 & 0.6456 & 0.9731 & 0.8913  \\
Yolov5n-CBi-v4-Distill-Yolov5m-CBi & 0.9391 & 0.65 & 0.9550 & 0.9077  \\
Yolov5n-CBi-v5-Distill-Yolov5m-CBi & 0.9378 & 0.6401 & 0.9593 & 0.8920  \\
\bottomrule
\end{tabularx}
\end{table*}

Table\ref{tab:val_dut} summarizes the comparison using four evaluation metrics: \textbf{mAP@0.5}, \textbf{mAP@0.5:0.95}, \textbf{Precision}, and \textbf{Recall}. Each competing model introduces architectural improvements to enhance drone detection, particularly for small targets.

\begin{itemize}
    \item \textbf{LA-YOLO}\cite{97}: It is based on the YOLOv5 baseline and introduces the SimAMC3 module, incorporating the SimAM attention mechanism to enhance feature extraction. It also combines the normalized Wasserstein distance loss with the CIoU loss to improve the sensitivity for small object detection.

    \item \textbf{SEB-YOLOv8s}\cite{99}: It is based on YOLOv8s and uses the SPD-Conv module to retain shallow semantic information and introduces the AttC2f module to better utilize spatial features. The integrated EMA module avoids downscaling, enhancing performance on low-quality images and images with small drones.

    \item \textbf{DRBD-YOLOv8s}\cite{98}: This model substitutes standard convolutions with depth-wise separable convolutions in the backbone and introduces the RCELAN module in the neck, which integrates feature segmentation and gradient path optimization. These changes improve multi-level feature extraction while reducing computational complexity.

    \item \textbf{IASL-YOLO}\cite{95}: It is a lightweight version of YOLOv8s, which replaces PAN-FPN with CFE-AFPN for progressive feature fusion and swaps the CIoU loss with the SIoU loss for better bounding box alignment. The LAMP pruning algorithm further compresses the model without a significant loss in accuracy.
\end{itemize}

\vspace{0.5em}
The proposed \textbf{YOLOv5-CBi-v2} model significantly outperforms the baseline YOLOv5 and several state-of-the-art drone detection methods on the DUT-AntiUAV~\cite{20} dataset, as shown in Table~\ref{tab:val_dut}. Compared to baseline YOLOv5, the proposed model achieves notable improvements across all evaluation metrics: mAP@0.5 increases from 0.9231 to \textbf{0.9545}, mAP@0.5:0.95 improves from 0.6274 to \textbf{0.6711}, precision increases from 0.9641 to \textbf{0.9716}, and recall improves from 0.8504 to \textbf{0.9262}. Furthermore, YOLOv5-CBi-v2 surpasses recent competitive models such as LA-YOLO, SEB-YOLOv8s, DRBD-YOLOv8s and IASL-YOLO by achieving the highest values in both mAP metrics and demonstrating a balance between precision and recall. These results clearly indicate that the proposed model delivers enhanced detection performance and robustness, particularly in challenging UAV detection scenarios.

An important observation from this study is that newer or more complex architectures are not necessarily better suited for all applications. Although most competing models are based on the latest YOLOv8 architecture, the proposed model is built on the relatively older YOLOv5 version. Despite this, it consistently outperforms the YOLOv8-based models in all key metrics. This supports a broader understanding that the effectiveness of a model depends on task-specific tuning and structural adaptation rather than adopting the latest versions. In applications like drone detection, where small object sensitivity and real-time reliability are crucial, a well-optimized earlier version can outperform more recent alternatives.

\section{Conclusion}\label{sec:conclusion}
This paper presents a comprehensive study on improving drone detection, with a specific focus on detecting small drones. A modified YolovN architecture (YolovN-CBi) is proposed that combines both CBAM and BiFPN to introduce attention on multiscale features without sacrificing inference speed. An extensive evaluation is performed on four public and a local test dataset for small object detection, demonstrating that Yolov5-CBi consistently outperforms other Yolo architectures (v8–v12) in both accuracy and generalization. Four variants (YolovN-CBi-v2, YolovN-CBi-v3,YolovN-CBi-v4,YolovN-CBi-v5) are proposed, with architectural changes based on the placement of the CBAM and BiFPN components in the backbone and neck of the Yolo architecture. Yolov5-CBi-v2 outperforms all variants and also reports better accuracy than state-of-the-art drone detection models, showing that early attention from initial layers improves small object detection.

In addition to architecture-level improvements, the challenge of deploying these models on edge devices is addressed by exploring knowledge distillation strategies. It is shown that adversarial distillation, especially when both the teacher and the student use the CBAM enhanced architecture, yields the best trade-off between detection performance and computational efficiency. The distilled Yolov5n-CBi student achieves high recall and precision with real-time inference capability, making it a viable solution for real-world surveillance applications where power and latency are constrained.

In conclusion, this study shows that carefully designed enhancements and compression strategies can yield lightweight but powerful models, better suited to real-world deployment than simply adopting the latest architectures.




\bibliographystyle{bst/sn-nature}
\bibliography{sn-bibliography}

\end{document}